\documentclass[10pt,twocolumn,letterpaper]{article}

\usepackage{cvpr}              

\usepackage[dvipsnames]{xcolor}
\newcommand{\red}[1]{{\color{red}#1}}

\definecolor{cvprblue}{rgb}{0.21,0.49,0.74}
\usepackage[pagebackref,breaklinks,colorlinks,citecolor=cvprblue]{hyperref}

\usepackage{multicol}
\usepackage{multirow}
\usepackage{amssymb}
\usepackage{amsmath}
\usepackage{wrapfig}
\usepackage{xspace}
\usepackage{color}
\usepackage{colortbl}
\usepackage[dvipsnames, svgnames, x11names, table]{}
\usepackage{xcolor}
\usepackage[misc]{ifsym}
\usepackage{makecell}
\usepackage{soul}
\usepackage{bm}
\usepackage{wrapfig}
\usepackage{subcaption, overpic, textpos}
\usepackage{ulem}
\usepackage{adjustbox}
\usepackage{pifont}
\usepackage{caption}
\usepackage{array}
\usepackage{enumitem}
\newcolumntype{C}[1]{>{\centering\arraybackslash}p{#1}} 

\newcommand{\cmark}{\ding{52}\xspace}%
\newcommand{\xmark}{\ding{56}\xspace}%

\definecolor{lightgray}{rgb}{0.8, 0.8, 0.8}
\definecolor{lgray}{rgb}{0.66, 0.66, 0.66}
\definecolor{whit_tab}{RGB}{255, 255, 255}
\definecolor{gray_tab}{RGB}{235, 235, 235}
\definecolor{oran_tab}{RGB}{254, 247, 241}
\definecolor{blue_tab}{RGB}{200, 227, 245}
\definecolor{lblu_tab}{RGB}{231, 239, 248}

\newcommand{\blue}{\textcolor[RGB]{0, 0, 255}}

\def\onedot{.\xspace}
\def\eg{\textit{e.g}\onedot} 
\def\Eg{\textit{E.g}\onedot}
\def\ie{\textit{i.e}\onedot}

\def\cf{\textit{c.f}\onedot}

\def\etc{\textit{etc}\onedot}

\def\etal{\textit{et al}\onedot}

\usepackage{lipsum}

\def\paperID{4403} 
\def\confName{CVPR}
\def\confYear{2024}

\title{Real-IAD: A Real-World Multi-View Dataset for Benchmarking Versatile Industrial Anomaly Detection}

\author{Chengjie Wang\textsuperscript{1,2} 
\quad Wenbing Zhu\textsuperscript{3,4} 
\quad Bin-Bin Gao\textsuperscript{2} 
\quad Zhenye Gan\textsuperscript{2} \\
\quad Jiangning Zhang\textsuperscript{2}
\quad Zhihao Gu\textsuperscript{1} 
\quad Shuguang Qian\textsuperscript{4} 
\quad Mingang Chen\textsuperscript{5} 
\quad Lizhuang Ma\textsuperscript{1}\thanks{Corresponding authors.} \\
\normalsize \textsuperscript{1}{Shanghai Jiao Tong University} \quad \textsuperscript{2}{Youtu Lab, Tencent} \quad \textsuperscript{3}{Fudan University}  \quad \textsuperscript{4}{Rongcheer Co., Ltd} \quad \\
\normalsize \textsuperscript{5}{Shanghai Development Center of Computer Software Technology} \\
{\tt\small Website: \url{https://realiad4ad.github.io/Real-IAD}}
}

\begin{document}
\maketitle
\begin{abstract}
Industrial anomaly detection (IAD) has garnered significant attention and experienced rapid development. However, the recent development of IAD approach has encountered certain difficulties due to dataset limitations. On the one hand, most of the state-of-the-art methods have achieved saturation (over 99\% in AUROC) on mainstream datasets such as MVTec, and the differences of methods cannot be well distinguished, leading to a significant gap between public datasets and actual application scenarios. On the other hand, the research on various new practical anomaly detection settings is limited by the scale of the dataset, posing a risk of overfitting in evaluation results. Therefore, we propose a large-scale, Real-world, and multi-view Industrial Anomaly Detection dataset, named Real-IAD, which contains 150K high-resolution images of 30 different objects, an order of magnitude larger than existing datasets. It has a larger range of defect area and ratio proportions, making it more challenging than previous datasets. To make the dataset closer to real application scenarios, we adopted a multi-view shooting method and proposed sample-level evaluation metrics. In addition, beyond the general unsupervised anomaly detection setting, we propose a new setting for Fully Unsupervised Industrial Anomaly Detection (FUIAD) based on the observation that the yield rate in industrial production is usually greater than 60\%, which has more practical application value. Finally, we report the results of popular IAD methods on the Real-IAD dataset, providing a highly challenging benchmark to promote the development of the IAD field.
\end{abstract}

\section{Introduction} \label{sec:intro}
High-quality datasets play a pivotal role in the development of computer vision technology, guiding the direction of technological advancement and bridging the gap between technical research and practical application. For instance, ImageNet~\cite{imagenet} has made an indelible contribution to the development of deep learning model structures and learning methodologies. MVTec AD~\cite{mvtec} abstracts defect detection in manufacturing into a research topic, thereby bringing visual learning algorithms closer to industrial production application scenarios. Nevertheless, the data volume of current AD datasets is still small and needs further development.

Industrial production is the cornerstone of human societal development, with product quality inspection playing a key role~\cite{cao2024survey}. Defects in the production process of parts can affect product quality and subsequently reduce product lifespan. In the production of pharmaceuticals, food, batteries, and other products, product defects can pose a threat to human safety. Given the importance of defect detection, manufacturing companies have invested substantial resources in this area~\cite{mvtec,mvtecloco,mvtec3d}, and academia has also gradually turned its attention to this issue in recent years~\cite{survey_ad,ganomaly,cflow,rd,simplenet,uniad}. Some automation technologies have begun to play a role in practical applications. Early applications of visual learning technology in industrial defect detection mainly involved supervised learning for detection~\cite{fastrcnn,fasterrcnn,cascadercnn} and segmentation~\cite{maskrcnn,deeplabv3,upernet} tasks. Although these methods have practical value, they still face technical challenges such as the need for precise defect location annotation. Moreover, the detection performance significantly decreases for defects that are scarce or not included in the training set. With the advent of datasets like MVTec AD~\cite{mvtec} and VisA~\cite{visa}, a large number of unsupervised anomaly detection methods have emerged~\cite{patchcore,rd,ocrgan,uniad,destseg,simplenet}. These techniques only need to ensure that the images in the training set are anomaly-free, and resulting models have the ability to predict defect locations and pixel areas, significantly reducing manual annotation costs and endowing the algorithm with the ability to recognize unknown defects.

The emergence of datasets like MVTec AD~\cite{mvtec}, and VisA~\cite{visa} has stimulated academic interest in industrial anomaly detection research and has given birth to many innovative methods. However, with technological progress, these datasets have gradually revealed their limitations. For example, the defect range in MVTec AD is small, and the actual application scenarios are simple. Recent methods have exceeded 99\% in I-AUROC (image-level) and P-AUROC (pixel-level) metrics, making it difficult to distinguish the merits of new methods. The scale of the data is also limited, with some categories in MVTec AD having only about 60 or even fewer defect images, and the random error caused by the small number of test images cannot be ignored. Some new IAD research uses some defective images in the training set, resulting in a smaller test set and exacerbating the problem. In terms of object diversity, MVTec AD includes 15 types of objects, and VisA includes 12 types of objects. The limited number of object types affects the evaluation of the capabilities of a unified model, leading to overly optimistic model metrics. In addition, there is a gap between the current dataset setup and the ideal application scenario. Although IAD is unsupervised learning, the training set still requires manual annotation. This process may introduce noise samples, such as the noise samples in the BTAD training set proposed in SoftPatch~\cite{softpatch}. Furthermore, mainstream 2D IAD datasets consist of single-view images, but in practical applications, part structures are complex, and a single view cannot cover all defects. Although some datasets, like MVTec AD-3D~\cite{mvtec3d}, attempt to solve the multi-view problem from a 3D perspective, the high cost of 3D sensors limits them in practical applications.

To make the dataset closer to actual application scenarios and address the limitations of existing datasets, we propose a new \textbf{Real}-world \textbf{I}ndustrial \textbf{A}nomaly \textbf{D}etection dataset called Real-IAD. This dataset far exceeds existing datasets in terms of the number of object categories and the number of images. To our knowledge, this dataset is the first to consider the multi-view problem in 2D IAD tasks. 
In an ideal scenario for practical application, image acquisition equipment is deployed on the production line, and the algorithm is trained automatically without human intervention, possessing the ability to determine whether a product is defective. 
Based on the characteristic that the yield rate of most production lines is greater than 60\%, we add 0$\sim$40\% of quality inspection anomaly images to the anomaly-free train set, proposing the completely unsupervised anomaly detection problem for the first time. Through this new dataset, we hope to promote technological development in the field of industrial anomaly detection, encourage the emergence of more efficient and practical detection methods, and provide stronger technical support for industrial production. In this new real-world dataset, we have collected 30 industrial products covering a variety of materials, such as plastic, wood, ceramics, and mixed materials \etc, considering the diversity in industrial production. We have also deliberately increased the resolution of the dataset to capture more subtle defect features, providing strong support for high-precision defect detection. In addition, we have collected images of each product from multiple angles to address the problem that a single view cannot cover all defects. To verify the effectiveness of the new dataset and promote technological development, we conducted a series of experiments on the dataset. Specifically, we conduct benchmark tests on existing unsupervised anomaly detection algorithms to assess their performance on the new dataset. The experimental results show that although these algorithms have achieved good performance on the original datasets, there is still room for improvement on the new dataset, indicating that the new dataset has a certain level of challenge, which can help promote algorithm improvement and innovation.

In summary, our main contributions are as follows:

\begin{itemize}
    \item[$\bullet$] 
We propose a new Real-IAD dataset, which is more than ten times larger than existing mainstream datasets. It includes 30 classes of objects with each containing 5 shooting angles, totaling 150K high-resolution images. Moreover, Real-IAD presents more challenging defects with a larger range of defect area and ratio proportions, better differing the performance between different methods and meeting various research settings of IAD.
    \item[$\bullet$] 
We construct a Fully Unsupervised IAD setting on the Real-IAD dataset that is closer to actual application scenarios, where only naturally existing constraint is used that the yield rate of most production lines is greater than 60\%, without introducing additional manual annotations.
    \item[$\bullet$] 
We report the performance of popular IAD methods on the Real-IAD dataset in several settings and provide a highly challenging benchmark to promote the development of the anomaly detection field.

\end{itemize}

\section{Related Work} \label{sec:rel}

\subsection{Anomaly Detection Datasets} \label{sec:rel_dataset}
Early anomaly detection works are generally conducted on KolektorSDD~\cite{KolektorSDD} which only contains one category and greatly limits the evaluation and development of algorithms. Subsequently, datasets such as MTD~\cite{mtd}, MPDD~\cite{mpdd} and BTAD~\cite{btad} are proposed, but these datasets are still relatively small, in terms of the number of categories and total number of images. Since the introduction of the MVTec AD dataset~\cite{mvtec}, the conventional sensory IAD tasks have started to get on track and gradually attracted a large number of researchers and practitioners. This dataset contains 15 industrial products in 2 types with a total of 5,354 images, greatly promoting algorithm research. Later VisA~\cite{visa} covers 12 objects in 3 types with a total of 10,821 images, elevating the number of IAD dataset to the 10K level for the first time while having a larger number of categories, \ie, 15. Recently, Zhou~\etal~\cite{pad} propose a synthetic PAD to promote research on pose-agnostic anomaly detection, but there is a natural gap between synthetic data and real samples, leading to inconsistencies in the metrics. In addition, some datasets have extended 3D information to better detect defects, \eg, MVTec 3D AD~\cite{mvtec3d}, Eyecandies~\cite{eycandies}, and Real3D~\cite{real3d}. Nevertheless, the current visual IAD datasets are still small in scale and category, generally applied to limited industrial scenarios, and there is no IAD dataset equivalent to the status of ImageNet-1K~\cite{imagenet} in the classification field. To address this issue, we propose a novel large-scale (approximately 150K), more categories (30 categories), and multi-view (5 shooting angles) Real-IAD dataset for the task, which contain high-definition images, elaborate labelings, and more challenging detects, committed to providing a more challenging benchmark to promote the development of IAD methods and fair comparison of different methods (\cf, \cref{sec:ben} for details).

\begin{figure*}[htp]
    \centering
    \includegraphics[width=1.0\linewidth]{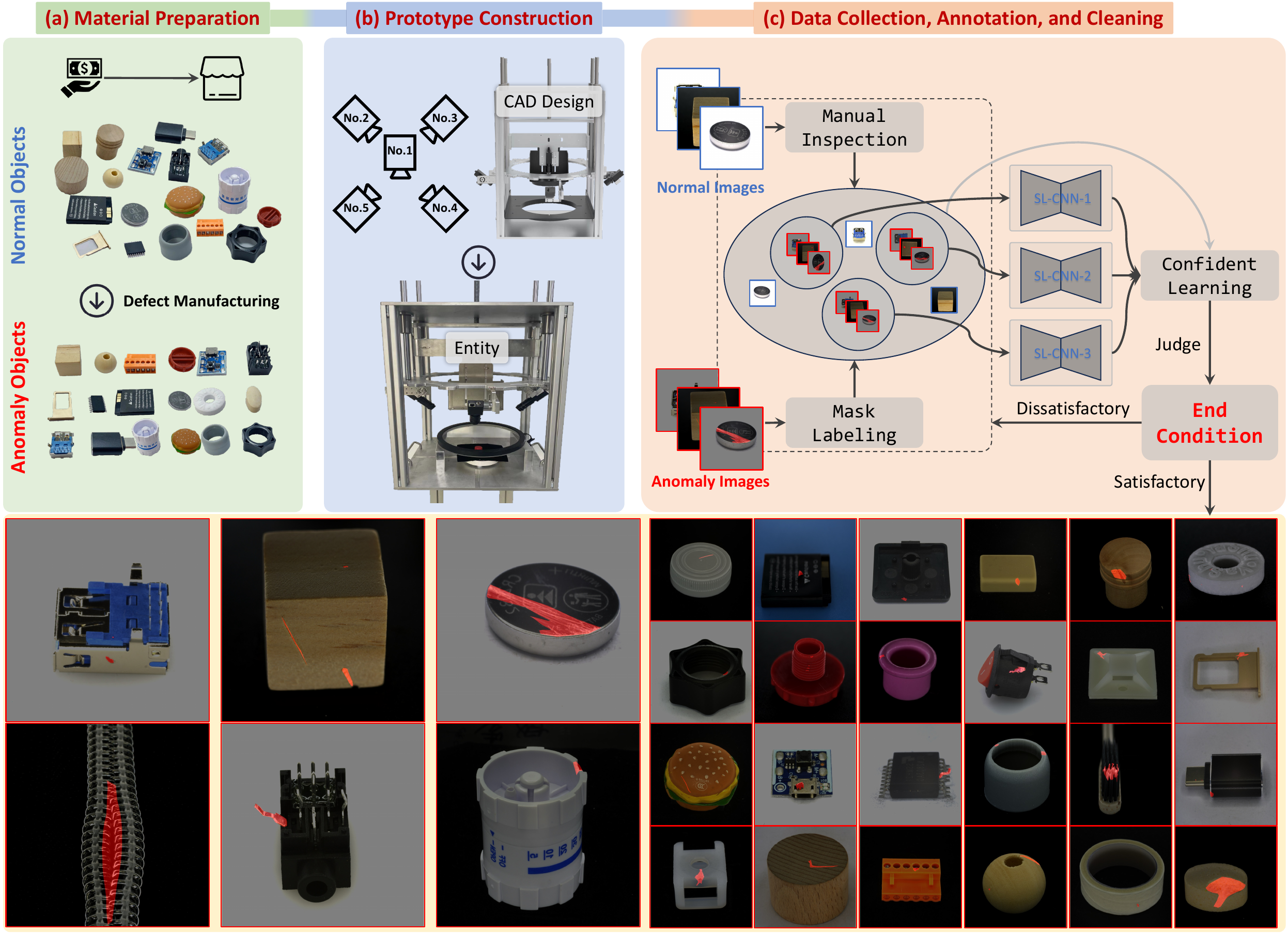}
    \caption{\textbf{Data collection pipeline for our proposed Real-IAD dataset,} which consists of four steps in tandem: \textbf{(a)} Material preparation and defect manufacturing. \textbf{(b)} Prototype design and construction that contains 5 shots for capturing multi-view images simultaneously, \ie, one top-down camera with extra four cameras arranged uniformly at 45 degrees. \textbf{(c)} Collection procedure that includes cyclical processes of data collection, annotation, and cleaning. 
    At the bottom, partial visualization of the final Real-IAD dataset reveals that the proposed Real-IAD exhibits large scale (30 classes), a wide range of defect proportions (0.01\% to 6.75\%), and a broad defect ratio (1:1 to 1:10), indicating that Real-IAD is highly challenging. Abnormal areas are prominently marked in red.
    }
    \label{fig:pipeline}
\end{figure*}

\subsection{Standard Anomaly Detection} \label{sec:rel_method}
The standard IAD task aims to identify whether a given image of a target class is anomalous or not and localize the anomaly region if the prediction is anomalous. The lack of anomalous data makes it more challenging and it is usually formulated as an unsupervised learning problem with only normal data available in the training phase.
Recently, vast efforts are dedicated to developing unsupervised anomaly detectors and form several mainstream approaches, namely data augmentation-based methods~\cite{cutpaste,draem}, reconstruction-based methods~\cite{da,ocrgan}, and embedding-based methods~\cite{patchcore,csflow,rd,simplenet}. Specially, the embedding-based methods can be further divided into four categories, \ie, memory bank~\cite{patchcore}, normalizing flow~\cite{csflow}, knowledge distillation~\cite{rd,zhang2023exploring}, and classification~\cite{simplenet}. These methods produce superior results.

\subsection{Other Settings in Anomaly Detection} \label{sec:rel_othersetting}
More challenging settings have been proposed one after another. 
The zero-/few-shot IAD~\cite{regad,aprilgan,clipad} focuses on employing a few normal samples for IAD, reducing the demand for data. 
SoftPatch~\cite{softpatch} introduces noisy data (less than 10\%) into the anomaly-free training data in the standard setting to model the real scenario, and Zaheer~\etal propose unsupervised video detection~\cite{zaheer2022generative}. 
Contrarily, the semi-supervised IAD~\cite{memseg} introduces anomalous data during training for better IAD. 
To avoid re-training for different categories, the unified IAD~\cite{uniad} accomplished IAD for multiple classes by a unified framework. In this paper, we propose a new setting called Fully Unsupervised Industrial Anomaly Detection (FUIAD) based on the observation that the yield rate in industrial production is usually greater than 60\%, which has a more practical application value. 
\section{Real-IAD Dataset Description} \label{sec:data}

\begin{table*}[htp]
    \centering
    \caption{\textbf{Comparison with current popular 2D real-world IAD datasets on different attributes.} \cmark: Satisfied. \xmark: Unsatisfied.}
    \label{tab:dataset}
    \renewcommand{\arraystretch}{1.2}
    \setlength\tabcolsep{3.0pt}
    \resizebox{1.0\linewidth}{!}{
        \begin{tabular}{p{1.7cm}<{\centering} p{1.2cm}<{\centering} p{1.5cm}<{\centering} p{1.5cm}<{\centering} p{1.5cm}<{\centering} p{1.5cm}<{\centering} p{2.5cm}<{\centering} p{2.5cm}<{\centering} p{2.5cm}<{\centering}}
            \toprule[0.17em]
            \multirow{2}{*}{Dataset} & \multirow{2}{*}{Class} & \multicolumn{3}{c}{Image Number} & \multirow{2}{*}{\makecell[c]{Image\\Resolution}} & \multirow{2}{*}{\makecell[c]{Defect Types \\ Per Category}} & \multirow{2}{*}{\makecell[c]{Segmentation \\ Labeling}} & \multirow{2}{*}{\makecell[c]{Multiple\\ Shooting Angles}} \\ 
            \cline{3-5}
            & & \multirow{1}{*}{Normal} & \multirow{1}{*}{Anomaly} & \multirow{1}{*}{All} & & & \\
            \hline
            BTAD        & 3  & 952   & 392   & 1344   & 600$\sim$1,600      & 3.00 & \cmark & \xmark \\
            SSGD        & 1  & 0     & 2504  & 2504   & 1,500$\times$1,000  & 7.00 & \xmark & \xmark \\
            MVTec AD    & 15 & 4,096 & 1,258 & 5,354  & 700$\sim$1,024      & 4.87 & \cmark & \xmark \\
            VisA        & 12 & 9,621 & 1,200 & 10,821 & 960$\sim$1,562    & 7.50 & \cmark & \xmark \\
            \hline
            Read-IAD    & 30 & 99,721 & 51,329 & 151,050  & 2,000$\sim$5,000  & 4.37 & \cmark & \cmark \\
            \toprule[0.12em]
        \end{tabular}
    }
\end{table*}

\begin{figure*}[htp]
    \centering
    \includegraphics[width=1.0\linewidth]{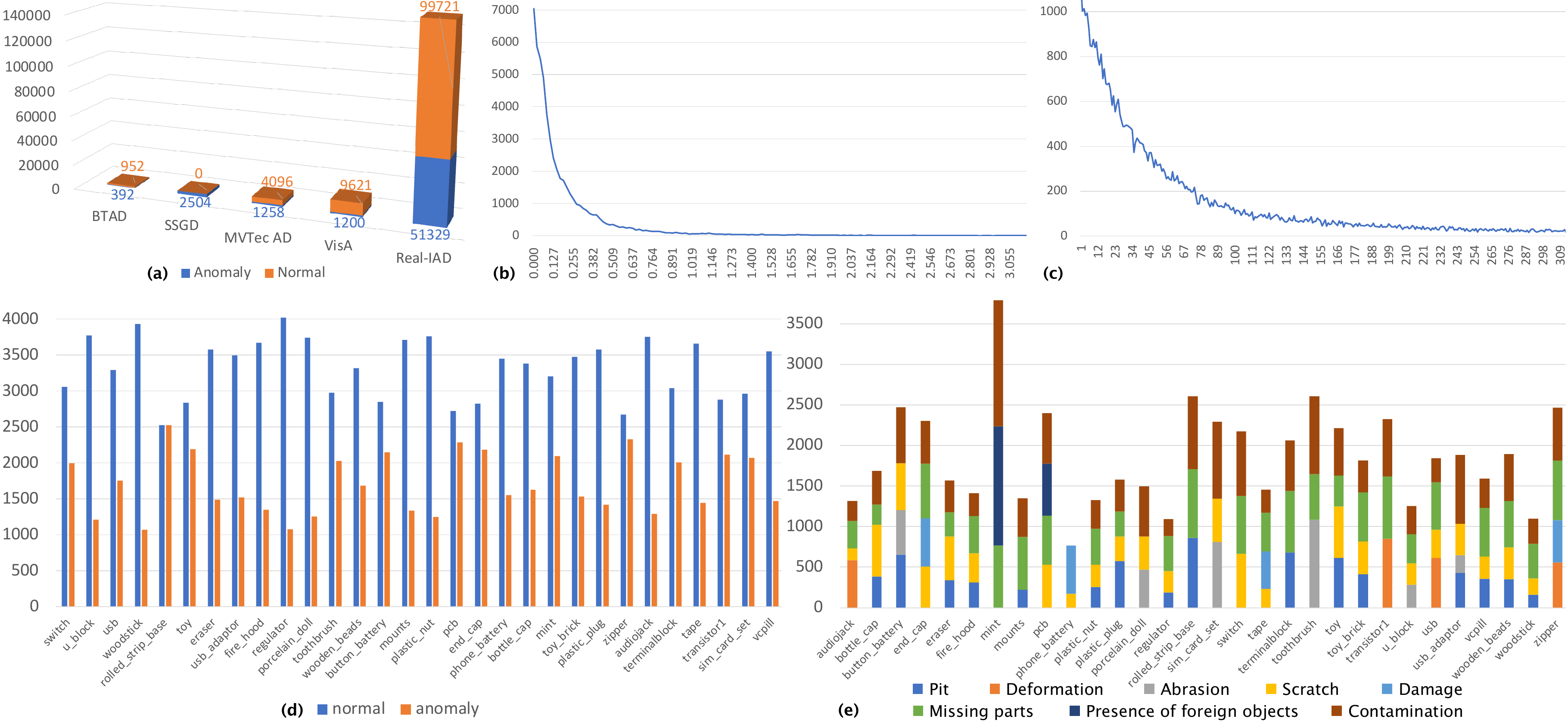}
    \caption{\textbf{Statistic information of our proposed Real-IAD datast.} \textbf{\textit{(a)}} Statistic comparison of anomaly/normal data volume among popular datasets. \textbf{\textit{(b)}} Statistics of the percentage of the image area occupied by the anomaly region. \textbf{\textit{(c)}} Statistics of the aspect ratio of the minimum bounding rectangle of the defect. \textbf{\textit{(d)}} Distribution of anomaly/normal image quantities across different categories. \textbf{\textit{(e)}} Distribution of data volume across different defect categories.
    }
    \label{fig:statistic}
    \vspace{-1.5em}
\end{figure*}

\begin{figure}[htp]
    \centering
    \includegraphics[width=1.0\linewidth]{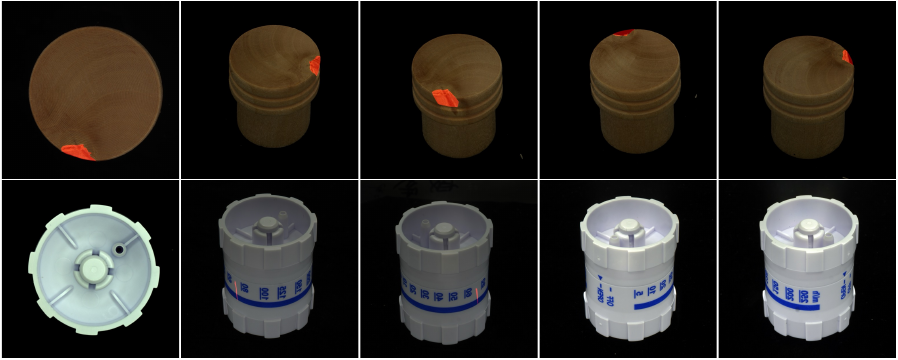}
    \caption{\textbf{Visualization of multi-view collections}, where the first column represents the results captured by the top-down camera, and the other columns represent the results captured by four surrounding cameras at 45 degrees. Abnormal images are highlighted for visualization, and normal images are displayed with normal brightness. It can be seen that multi-view shooting can solve the problem of invisibility of abnormal areas from a single viewpoint. 
    }
    \label{fig:vis_multi_view}
    \vspace{-1.5em}
\end{figure}

\subsection{Data Collection and Construction Manner} \label{sec_data:collect_manner}
This section introduces the construction pipeline of Real-IAD in \cref{fig:pipeline}, and the procedure includes three parts:

\noindent\textbf{\textit{1)} Material Preparation.} 
We have purchased and collected 30 objects covering a variety of materials, including metal, plastic, wood, ceramics, and mixed materials. Some objects are shown in the upper part of \cref{fig:pipeline}-(a). We manually created various types of defects, including missing parts, dirt, deformation, pits, damage, holes, cracks, scratches, as shown in the lower part of \cref{fig:pipeline}-(a). Subsequently, these normal and anomaly objects are sent to the prototype machine for image collection.

\noindent\textbf{\textit{2)} Prototype Construction.} The construction of the acquisition equipment is shown in \cref{fig:pipeline}-(b). There are a total of five cameras capturing the object from different angles, with one camera taking top shots and the other four taking shots from symmetric angles at approximately 45 degrees. Additionally, to better image the object and clearly capture minor defects, a ring light source is installed above the object. In actual applications, automated equipment is used to flip the parts for bottom quality inspection. For parts with more structural obstructions, the camera is rotated to capture images from more angles. Considering that part flipping and additional shooting angles may affect actual applications but not algorithm research, we abstract the multi-view visual quality inspection problem in real scenarios into five shooting angles. The camera adopted is HIKROBOT MV-CE200-10GC with 3,648$\times$5,472 resolution.

\noindent\textbf{\textit{3)} Data Collection, Annotation, and Cleaning.}
As shown in\cref{fig:pipeline}-(c), to ensure the accuracy of dataset collection, we first manually confirm all normal images and anomaly images and use LabelMe to manually label pixel-level masks for anomaly data. Then, the data is divided into three groups, each of which is supervised trained using a cascade RCNN based on HRNet-32w backbone for confident learning. We look for inconsistencies between the model's prediction results and the manual annotation results. These data will undergo manual inspection and annotation again until the model's predicted AP results remain essentially unchanged, and the number of annotated images that need to be modified is less than a certain threshold. Finally, we consider the data to be clean and construct different dataset settings.

\subsection{Comparison with Popular 2D Datasets} \label{sec_data:comparison}

This section provides a detailed statistical analysis of the proposed Real-IAD dataset, and compares it with mainstream datasets, \ie, MVTec AD~\cite{mvtec} and VisA~\cite{visa}.

\noindent\textbf{Comparison with Mainstream Datasets.}
As shown in \cref{tab:dataset}, compared to mainstream datasets, our method has at least a 2x increase in the number of categories, and a magnitude increase in quantity, \ie, from 10K to 150K. Furthermore, Real-IAD provides images with a resolution of up to 2,000$\sim$5,000 to support research on finer-grained IAD algorithms, which involves cropping from original images. Furthermore, each object is provided with five images taken from different angles with segmentation labeling to support a multi-view setting, \cf, \cref{fig:vis_multi_view}.

\noindent\textbf{Data Statistics.}
\cref{fig:statistic} shows statistic of our Real-IAD dataset. Compared to the contrastive datasets, our Real-IAD has an order of magnitude improvement on both normal and abnormal data (\cref{fig:statistic}-a). At the same time, the proportion of defect area (\cref{fig:statistic}-b) and the range of defect ratio (\cref{fig:statistic}-c) are larger, indicating a higher difficulty level of the dataset, which is also proven by the experiments in \cref{tab:uad-realiad}. The bottom of \cref{fig:statistic} displays the number of normal and abnormal data in different categories (\cref{fig:statistic}-d), as well as the proportion of different types of defects (\cref{fig:statistic}-e). Overall, the ratio of normal to abnormal data is not very disparate, \eg, VisA reaches 8:1, and each type of category has multiple types of defects. 

\noindent\textbf{Advantage Analyses.}
Our Real-IAD dataset has advantages in the following aspects: \textbf{\textit{1)} Diversity}: Compared with existing datasets, the Real-IAD dataset covers a wider range of categories and provides richer scenarios, which helps to train more robust anomaly detection models and conduct fair evaluations. \textbf{\textit{2)} Large-scale}: Real-IAD is the first to provide a dataset of over 150K images, a magnitude increase compared to popular IAD datasets, which also provides multi-view images with pixel-level annotations. \textbf{\textit{3)} Challenge}: Compared with existing datasets, the Real-IAD has a higher level of difficulty, which can drive the development and progress of current anomaly detection algorithms.

\subsection{Real-IAD Visualization} \label{sec_data:vis}
The bottom of \cref{fig:pipeline} displays 30 sample data types from Real-IAD, demonstrating that the dataset includes data from a variety of material types, such as metal, plastic, wood, ceramics, and mixed material. Moreover, the types of defects include pit, deformation, abrasion, scratches, damage, missing parts, foreign objects, and contamination. The defect area occupies a proportion 
ranging from 0.1 to 0.5, with a ratio ranging from 0.1 to 10.0. The richer data indicates that Real-IAD is highly challenging, and it is expected to promote further development in IAD.

\section{Benchmark} \label{sec:ben}

\subsection{Evalution settings}
We establish two evaluation protocols, including Unsupervised IAD, and fully Unsupervised IAD for all 30 categories in Real-IAD dataset.

\noindent\textbf{Unsupervised IAD (UIAD):}
The algorithm learns patterns and structures only from normal samples. Therefore, most existing anomaly datasets are usually divided into training and testing sets, where the training set only contains normal samples. In contrast, the testing set consists of both normal and anomalous samples. Following the previous arts, we fisrt evaluate our Real-IAD with the UIAD setting.  

\noindent\textbf{Fully Unsupervised IAD (FUIAD):}
The setting is inevitable in real-world IAD but is seldom discussed. And it is difficult to construct the FUIAD setting due to the limited availability of anomalous samples in the existing datasets, such as MVTec AD~\cite{mvtec} and VisA~\cite{visa}. This paper is the first to consider FUIAD (\textit{i.e.}, UIAD with noise) from a dataset construction perspective. In the Real-IAD, large and diverse samples of anomalies give us the flexibility to build an FUIAD setup.

To create an FUIAD setting, we first need to fix the testing set including normal and anomaly samples. In our experiment, the number of normal and anomaly samples is uniformly set to 100 samples (500 images). The remaining normal and anomaly samples will be a candidate set for constructing a noisy training set. We keep the total number of samples in the training set constant but the corresponding numbers of normal and abnormal ones are adaptively adjusted with a given noise ratio. To do this, we need first infer the number of training samples according to the scale of the candidate set and the range of noisy ratio~(e.g., [0.1, 0.4]). Then, we randomly sample normal and anomaly samples from the candidate set to construct a training set with a specific noisy ratio. In this way, we obtain several new 
fully unsupervised benchmarks with different noise ratios dubbed $\alpha \in [0,1]$.

\subsection{Evaluation Metric}
The AUROC is the most widely used metric for image-level and pixel-level anomaly detection. 
In addition, a normalized Per-Region Overlap (PRO) between segmentation and ground truth is calculated and the Area Under PRO curve (AUPRO) is also adopted as a pixel-level metric
~\cite{mvtec}.

Most IAD methods~\cite{rd,patchcore,cutpaste,draem,destseg,csflow,simplenet} evaluate anomaly detection only at {image and pixel levels} due to limitations of existing IAD datasets. The proposed Real-IAD, to the best of our knowledge, is the first multi-view anomaly detection dataset, where each sample consists of multiple different views. Therefore, in addition to evaluating anomaly detection performance at the image and pixel levels, we integrate the results from multiple views to assess sample-level performance. This is more consistent with the indicator evaluation in industrial production lines.

\section{Comparisons with IAD Benchmarks}

\subsection{Results on Unsupervised IAD}

\begin{table*}[th]
    \centering
    \caption{UIAD performance comparisons on Real-IAD and some existing IAD datasets, MVTec AD~\cite{mvtec} and VisA~\cite{visa}. The results are averaged over all categories on each dataset. We report the averaged value and standard deviation (denoted as ``mean$\pm$std") based on the results of different methods for each metric. The lower the value means the corresponding anomaly detection task is more challenging.}
    \renewcommand{\arraystretch}{1.2}
    \resizebox{\linewidth}{!}{
        \begin{tabular}{l|l|ccc|cc|cc|c}
            \hline
            &  & \multicolumn{3}{c|}{Embedding-based}  & \multicolumn{2}{c|}{Data-Aug-based} & \multicolumn{2}{c|}{Reconstruction-based}  &\multirow{2}{*}{\textbf{Mean$\pm$Std}} \\ 
             \multirow{-2}{*}{\textbf{Datasets}}     & \multirow{-2}{*}{\textbf{Metric$\uparrow$}} 
            &\textbf{PatchCore\cite{patchcore}} & \textbf{PaDim\cite{padim}} & \textbf{CFlow\cite{csflow}}  & \textbf{SimpleNet\cite{simplenet}} & \textbf{DeSTSeg\cite{destseg}}  & \textbf{RD\cite{rd}} & \textbf{UniAD\cite{uniad}}   & \\ 
            \hline
            \rowcolor{blue!6}  \cellcolor{White}{} & I-AUROC   &99.0	&95.5	&97.1	&99.3	&98.4	&98.5	&97.5	&97.9	$\pm$1.2\\
            \multirow{-2}{*}{MVTec AD~\cite{mvtec}}                          
            & P-AUPRO   &94.7	&91.3	&92.3	&89.9	&94.7	&93.1	&90.7	&92.4	$\pm$1.8\\
            \hline
            \rowcolor{blue!6}  \cellcolor{White}{} & I-AUROC   &95.1	&85.1	&91.8	&95.4	&90.3	&95.9	&88.8	&91.8	$\pm$3.7\\
            \multirow{-2}{*}{VisA~\cite{visa}}                          
            & P-AUPRO   &91.2	&79.7	&84.8	&88.7	&90.6	&92.9	&85.5	&87.6	$\pm$4.2\\
            \hline
            \rowcolor{blue!6}  \cellcolor{White}{} & I-AUROC   &90.4	&86.6	&85.0	&91.7	&89.2	&87.6	&82.7	&87.6	$\pm$2.9\\
            \multirow{-2}{*}{Real-IAD (single-view)}                          
            & P-AUPRO    &92.5	&90.0	&88.6	&88.9	&90.3	&94.4	&86.0	&90.0	$\pm$2.6\\
            \hline
            & S-AUROC  &93.7	&91.2	&89.8	&94.9	&94.0	&83.7	&88.1	&90.3	$\pm$4.0\\
            \rowcolor{blue!6}  \cellcolor{White}{} & I-AUROC  &89.4	&80.3	&82.5	&88.5	&86.9	&87.1	&82.9	&85.0	$\pm$3.6\\
            \multirow{-3}{*}{Real-IAD (multi-view)}                          
            & P-AUPRO  &91.5	&81.8	&90.5	&84.6	&81.5	&93.8	&86.1	&86.3	$\pm$4.4\\
            \hline
        \end{tabular}
    }
    \vspace{-1.5em}
    \label{tab:uad-realiad}
\end{table*}

We comprehensively compare the performance of our Real-IAD with some popular IAD benchmarks (e.g., MVTec AD~\cite{mvtec} and VisA~\cite{visa}). MVTec AD~\cite{mvtec} is a widely used dataset in the field of industrial anomaly detection. It consists of high-resolution images of 10 objects and 5 textures, captured under different lighting conditions and with different types of anomalies.  
The dataset provides ground truth annotations for the location and type of anomalies present in the images, enabling quantitative evaluation of detection performance. VisA~\cite{visa} is 2$\times$ larger than MVTec AD, with both image and pixel-level annotations. It spans 12 objects across 3 domains, with challenging scenarios including complex structures in objects, multiple instances, and object pose/location variations. The proposed Real-IAD is a large-scale (approximately 150K) with more categories (30 objects), and multi-view (5 shooting angles) dataset for anomaly detection, where each anomaly image is labeled by a pixel-level mask and a specific defect type. Considering that MVTec and VisA are single-view, we also select one viewpoint (\textit{i.e.}, top view) from Real-IAD to form a single-view Real-IAD, a subset of multi-view Real-IAD.

For the UIAD setting, we mainly choose embedded-based IAD methods, such as PatchCore\cite{patchcore}, PaDim~\cite{padim}, and CFlow~\cite{csflow}, data-augmentation-based methods, SimpleNet~\cite{simplenet} and DeSTSeg~\cite{destseg}, and reconstruction-based RD~\cite{rd} and UniAD~\cite{uniad} for performance comparisons. PatchCore~\cite{patchcore} first extracts neighborhood-aware patch-level features and then stores them in a memory bank. At test time, images are classified as anomalies if at least one patch is anomalous, and pixel-level anomaly segmentation is generated by scoring each patch feature. PaDiM~\cite{padim} extract pre-trained features to model normal distribution, then utilize a distance metric to measure the anomalies.
CFlow~\cite{csflow} proposed to use a conditional normalizing flow framework to estimate the exact data likelihoods which are infeasible in other generative models for IAD. SimpleNet~\cite{simplenet} and DeSTSeg~\cite{destseg} convert Unsupervised IAD into supervised IAD training with generated anomaly images/features and real normal images/features. Feature reconstruction-based RD~\cite{rd} and UniAD~\cite{uniad} prevent the model from learning the shortcut with reverse distillation and neighborhood-masked attention, respectively. We reproduce PatchCore~\cite{patchcore}, PaDim~\cite{padim}, CFlow~\cite{csflow} and RD~\cite{rd} with open-sourced Anomalib, and other methods including SimpleNet~\cite{simplenet}, DeSTSeg~\cite{destseg} and UniAD~\cite{uniad} with official codes. In our experiments, we resize all images to 256$\times$256, only center crop 224$\times$224 from the resized 256$\times$256 for PatchCore~\cite{patchcore} and PaDim~\cite{padim}. Other hyperparameters, such as batch size and learning rate, are kept the same as the official implementation.

The results of all methods on MVTec, VisA, and our single-/multi-view Real-IAD are presented in Table~\ref{tab:uad-realiad}. 1) We can observe that there is a significant performance decrease from MVTec (97.9\% in I-AUROC) to our single-view and multi-view Real-IAD~(85\% in I-AUROC). This indicates that the proposed Real-IAD is more challenging than the existing datasets for anomaly detection. The performance drop is more pronounced when using a unified model, which is reasonable considering the more complex data distribution~(\textit{i.e.}, more views and more classes) in Real-IAD; 2) It is hard to evaluate different methods on the existing datasets as the results are very similar. Especially on MVTec, most methods achieve about 98\%-99\% in I-AUROC. In contrast, on the Real-IAD dataset, most methods only obtain about 90\% I-AUROC, which is better for evaluating the effectiveness of anomaly detection algorithms; 3) The pixel-level P-PRO on Real-IAD is comparable to the existing VisA but noticeably lower than MVTec AD. This indicates that the proposed Real-IAD dataset also presents challenges at the pixel-level anomaly localization.

\subsection{IAD Benchmarks Meet FUIAD}

Traditional Unsupervised IAD methods indeed assume that the training dataset only contains normal samples. However, in practical applications, it isn't easy to guarantee that all training samples are normal. Considering that the yield rate of good products on a production line is typically higher than 60\%, it means that no more than 40\% products are abnormal. Therefore, existing unsupervised methods may not be suitable for practical applications because they require manual annotation before training that is laborious.

Hence, a more practical solution is to perform fully unsupervised anomaly detection, which allows a certain proportion of abnormal samples in the training process. This paradigm can better adapt to real-world scenarios and reduce reliance on manual annotation. By building a fully Unsupervised IAD setting, it is possible to better simulate abnormal situations in real-world scenarios, thereby improving the robustness and practicality of IAD algorithms.

\begin{table}[t]
\centering
\caption{Comparison with popular IAD datasets on FUIAD setting. NR~($\alpha$) and \# TS are the noisy ratio and anomalous testing samples, respectively. The fail settings are marked in blue.}
\renewcommand{\arraystretch}{1.0}
\resizebox{\linewidth}{!}{
\begin{tabular}{lc|ccc|cccc}
\hline
       & & \multicolumn{3}{c|}{\textbf{\# valid category}}  & \multicolumn{4}{c}{\textbf{\# training images or samples}}    \\ 
  \cmidrule{3-9}
    \multirow{-2}{*}{\textbf{NR}} &\multirow{-2}{*}{\textbf{\# TS}}
  &{MVTec AD} &{ViSA} &{Real-IAD} &{MVTec AD} &{ViSA} &{Real-IAD(I)} & {Real-IAD(S)} \\
  \hline
  \multirow{4}{*}{$\alpha=0.1$} 
&25	&15	&12	&30	&268	&667	&2605	&521 \\
&50	&\blue{13}	&12	&30	&\blue{223}	&500	&2466	&493 \\
&75	&\blue{9}	&12	&30	&\blue{200}	&250	&2327	&465 \\
&100	&\blue{4}	&\blue{0}	&30	&\blue{181}	&\blue{0}	&2188	&437 \\
&150	&\blue{0}	&\blue{0}	&30	&\blue{0}	&\blue{0}	&1910	&382  \\
  \hline
  \multirow{4}{*}{$\alpha=0.2$} 
&25	&15	&12	&30	&252	&375	&2931	&586 \\
&50	&\blue{13}	&12	&30	&\blue{193}	&250	&2774	&554 \\
&75	&\blue{9}	&12	&30	&\blue{148}	&125	&2618	&523 \\
&100	&\blue{4}	&\blue{0}	&30	&\blue{120}	&\blue{0}	&2462	&492 \\
&150	&\blue{0}	&\blue{0}	&30	&\blue{0}	&\blue{0}	&2149	&429 \\
  \hline
  \multirow{4}{*}{$\alpha=0.4$} 
&25	&15	&12	&30	&152	&187	&3908	&781\\
&50	&\blue{13}	&12	&30	&\blue{108}	&125	&3699	&739\\
&75	&\blue{9}	&12	&30	&\blue{78}	&62	&3491	&698 \\
&100	&\blue{4}	&\blue{0}	&30	&\blue{60}	&\blue{0}	&3283	&656 \\
&150	&\blue{0}	&\blue{0}	&30	&\blue{0}	&\blue{0}	&2866	&573 \\
  \hline
\end{tabular}}
\vspace{-1.5em}
\label{tab:fuad-comrealiad}
\end{table}

\begin{table*}[th]
    \centering
    \caption{FUIAD performance (S-AUROC/I-AUROC/P-PRO) comparisons on Real-IAD. The results are averaged over 30 categories. The best results are marked in red and the second-best ones are marked in blue.}
    \renewcommand{\arraystretch}{1.2}
    \resizebox{1.0\linewidth}{!}{
        \begin{tabular}{l| cc cc| cc |cc}
        \hline
        \multirow{2}{*}{\textbf{Settings}} & \multicolumn{4}{c|}{Embedding-based}  & \multicolumn{2}{c|}{Data-Aug-based} & \multicolumn{2}{c}{Reconstruction-based} \\ 
        & \textbf{PaDim\cite{padim}} &\textbf{CFlow\cite{cflow}} &\textbf{PatchCore\cite{patchcore}}   &\textbf{SoftPatch\cite{softpatch}} & \textbf{SimpleNet\cite{simplenet}} & \textbf{DeSTSeg\cite{destseg}}  & \textbf{RD\cite{rd}} & \textbf{UniAD\cite{uniad}}  \\
        \hline                     
        $\alpha=0.0$ &92.7	/ 84.6	/ 84.4	&88.8	/ 83.9	/ 90.6	&94.4	/ \blue{91.3}	/ \blue{92.6}	&93.9	/ \red{91.4}	/ 92.1	&\red{94.9}	/ 89.8	/ 83.9	&\blue{94.6}	/ 89.6	 / 88.7	&85.5	/ 89.3	/ \red{95.0}	&89.6	/ 85.4	/ 87.6\\
        $\alpha=0.1$ &85.5	/ 81.9	/ 86.4	&81.8	/ 80.3	/ 90.7	&\blue{92.2}	/ \blue{90.4}	/ \blue{93.2}	&\red{92.8}	/ \red{90.9}	/ 92.9	&87.8	/ 83.3	/ 79.9	&88.1	/ 85.6	/ 86.9	&83.5	/ 88.1	/ \red{95.1}	&87.2	/ 84.2	/ 87.7\\
        $\alpha=0.2$ &82.2		/ 80.1		/ 86.5		&81.3	/ 79.6	/ 90.7	&\blue{91.1}		/ \blue{89.5}		/ \blue{93.0}		&\red{92.2}		/ \red{90.5}		/ 92.9	    	& 82.7		/ 79.6		/ 75.9		& 81.4		/ 80.3		/ 83.2		&80.7		/ 87.3		/ \red{94.9}		&85.2	/ 82.8                      	/ 87.3\\               
        $\alpha=0.4$ &76.5		/ 77.0		/ 86.1		&78.6	/ 78.0	/ 90.2	&\blue{88.9}		/ \blue{88.1}		/ {92.4}	        	&\red{90.6}		/ \red{89.3}		/ \blue{92.5}	&77.7		/ 74.7		/ 70.5		& 74.5		/ 74.4		/ 75.5		&76.8		/ 84.5		/ \red{94.7}		&81.5	/ 80.1                  	/ 86.6\\
        \hline
        \end{tabular}
    }
    \vspace{-1.5em}
    \label{tab:fuad-realiad}
\end{table*}

Popular anomaly detection datasets like MVTec AD~\cite{mvtec}
and VisA~\cite{visa} are primarily designed for unsupervised settings. To evaluate the capabilities of fully unsupervised algorithms, researchers have to randomly sample a portion of anomaly samples from the testing set and add them as noise samples to the training set, creating a fully unsupervised IAD experiment.  In this way, it leads to a significant reduction in the number of testing samples, which may be insufficient to effectively evaluate the performance of anomaly detection algorithms. For quantitative analysis, we assume a certain number of testing samples (e.g., 25, 50, 75, 100 and 150) for each category and calculate the number of valid testing categories and the number of training images or samples based on a given noise ratio (e.g., 0.1, 0.2 and 0.4). The corresponding statistic results are reported in Table~\ref{tab:fuad-comrealiad}.

We can see that increasing the noisy ratio or injecting anomalous noisy samples into the normal training set (MVTec and VisA) would lead to a decrease in the number of valid testing categories and the number of normal training samples, making it impossible to evaluate the performance of FUIAD algorithms in some challenging settings (high noisy ratios).
In contrast, the proposed Real-IAD ensures a consistent number of valid testing categories and a stable scale of corresponding normal samples in the training set when the noisy ratio ranges from 0.1 to 0.4 and the number of anomalous testing samples varies from 50 to 150. We believe that Real-IAD provides a more reliable and comprehensive assessment for FUIAD algorithms.

\section{Comparisons with Fully Unsupervised IAD}

For fully unsupervised anomaly detection, we set noisy ratio $\alpha \in \{0.1, 0.2, 0.4\}$. In addition, to make a fair comparison between FUIAD and UIAD settings on the same testing set, we also set the noisy ratio to 0, which transforms FUIAD into UIAD. The results of FUIAD ($\alpha=0$) should be the upper bound of all FUIAD settings ($\alpha>0$). Considering that it is not easy to establish an effective fully unsupervised setting on existing datasets, we only evaluate fully unsupervised anomaly detection methods on our Real-IAD. Similar to the evaluation of unsupervised anomaly detection, we also select the same methods, PatchCore\cite{patchcore}, PaDim~\cite{padim}, CFlow~\cite{csflow}, SimpleNet~\cite{simplenet}, DeSTSeg~\cite{destseg}, RD~\cite{rd} and UniAD~\cite{uniad}. To alleviate the effect of noise samples, SoftPatch~\cite{softpatch} proposes a denoising mechanism based on PatchCore for memory bank construction. Therefore, we also evaluate the SoftPatch method under the FUIAD setting on our Real-IAD. The main results are reported in Tabel~\ref{tab:fuad-realiad}. 

Under the setting of unsupervised anomaly detection (the noisy ratio is set to zero), most state-of-the-art anomaly detection methods (such as PatchCore~\cite{patchcore}, SoftPatch~\cite{softpatch}, SimpleNet~\cite{simplenet}, DeSTSeg~\cite{destseg}) show almost similar performance for sample- and image-level anomaly classification. For pixel-level anomaly segmentation, PatchCore~\cite{patchcore}, SoftPatch~\cite{softpatch} and RD~\cite{rd} that use multi-level features are more advantageous, thanks to the low-level features retaining rich spatial location information. 

In the setting of fully unsupervised anomaly detection, almost all methods suffer severe performance degradation on all metrics, especially PaDim~\cite{padim}, SimpleNet~\cite{simplenet}, DeSTSeg~\cite{destseg} and RD~\cite{rd}. Unsupervised PatchCore~\cite{patchcore} shows robustness, thanks to the patch-level Memory-Bank mechanism. In the Memory-Bank, normal and abnormal features are stored simultaneously, and stable detection can be achieved as long as the distribution of abnormal features during the inference phase is different from the stored anomalies. In addition, the proportion of abnormal pixels is actually very low even with a high noise ratio at sample-level (such as 0.4), and thus the quality of normal patch features in Memory-Bank can be ensured to some extent. The SoftPatch~\cite{softpatch} that first filters some noisy features and then constructs Memory-Bank achieves almost optimal performance in all methods. However, the performance improvement is relatively limited compared to unsupervised PatchCore~\cite{patchcore}. This means that fully unsupervised anomaly detection still requires more in-depth research efforts. \Eg, using model ensemble methods to improve robustness to noise, combining the semantic understanding capabilities of large vision language models to provide prior distribution of noisy data, \textit{etc}. 
\section{Conclusion} \label{sec:con}
Based on the analysis of existing anomaly detection datasets and practical industry applications, we have several observations. 
First of all, unsupervised industrial anomaly detection algorithms have almost reached saturation in performance, but it is still hard to deploy them in actual industrial inspection applications. 
Secondly, there is a gap between algorithms and applications, where most algorithms rely on clean normal training samples, but the data obtained in actual industrial production lines contains a certain amount of noise. 
Finally, fully unsupervised industrial anomaly detection is a more suitable setting for practical applications, but existing datasets are not sufficient to support research due to limited samples.
In order to address the above problems, we propose a large-scale, real-world,  multi-view anomaly detection dataset (Real-IAD), which contains 150k high-resolution images, covering 30 objects of metal, plastic, wood, ceramics, and mixed materials, each object contains 5 different views with 8 common defects. The Real-IAD provides high-quality annotations at different levels such as pixel, image, and sample. We established two settings, unsupervised and fully unsupervised, to conduct extensive and comprehensive evaluations using state-of-the-art anomaly detection methods. We hope that the Real-IAD can advance research in the field of anomaly detection.

\noindent\textbf{Limitation and Future Works.} This paper only reports the results of some typical methods on Real-IAD. In the future, we will replicate more methods on Real-IAD and provide results under more settings, \eg, zero-shot, few-shot, and semi-supervised settings, \etc In addition, considering the large-scale and multi-view characteristics of Real-IAD, it is also worthwhile to further study algorithms that are well-suited to these features.

\normalem
{
    \small
    \bibliographystyle{ieeenat_fullname}
    \bibliography{main}
}


\clearpage
\setcounter{page}{1}

\begin{figure*}[htbp]
  \centering
  \subcaptionbox{\label{fig:subfig1}}[0.95\linewidth]{
    \includegraphics[width=0.95\linewidth]{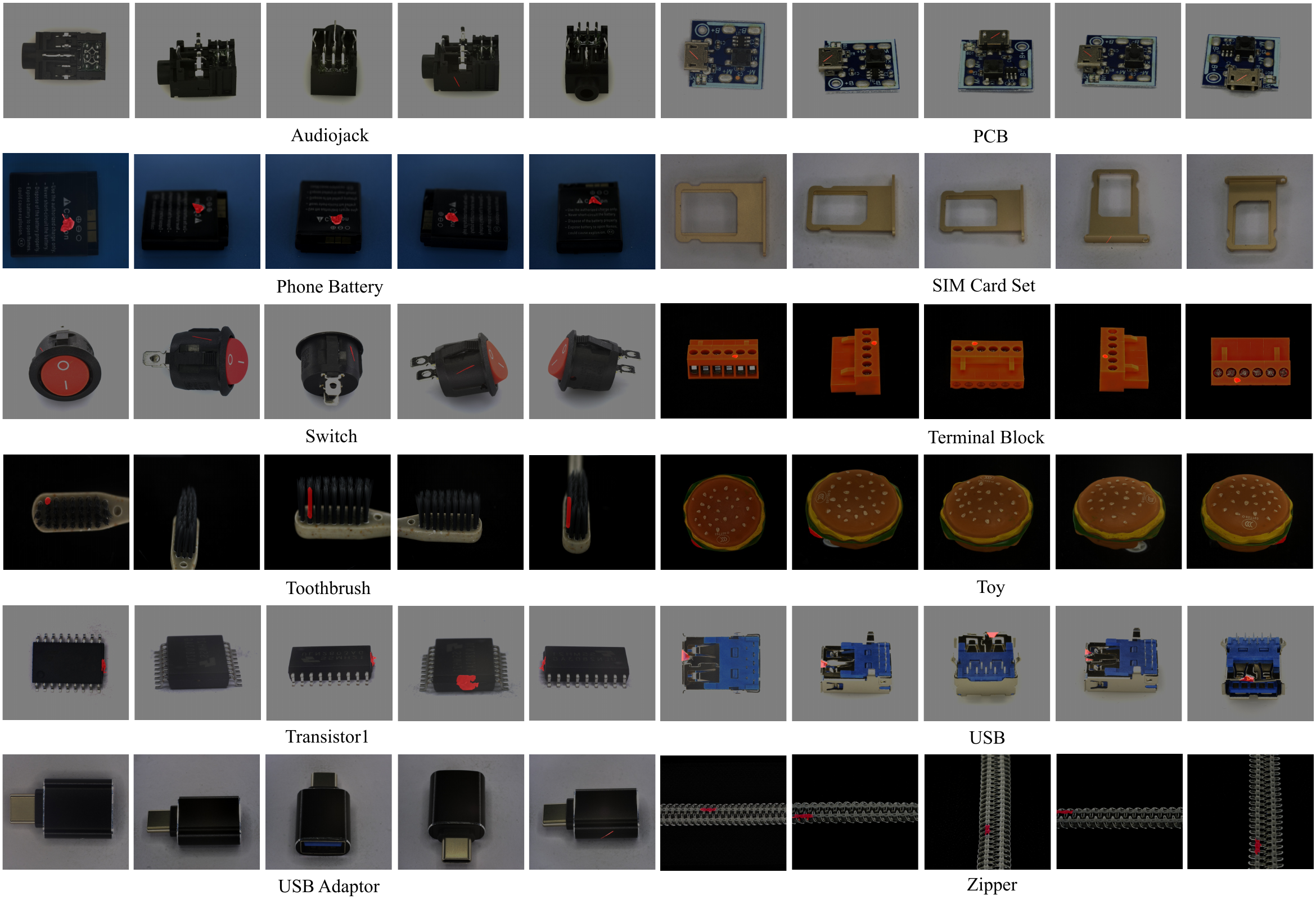}
  }
  \vspace{10pt}
  \subcaptionbox{\label{fig:subfig2}}[0.95\linewidth]{
    \includegraphics[width=0.95\linewidth]{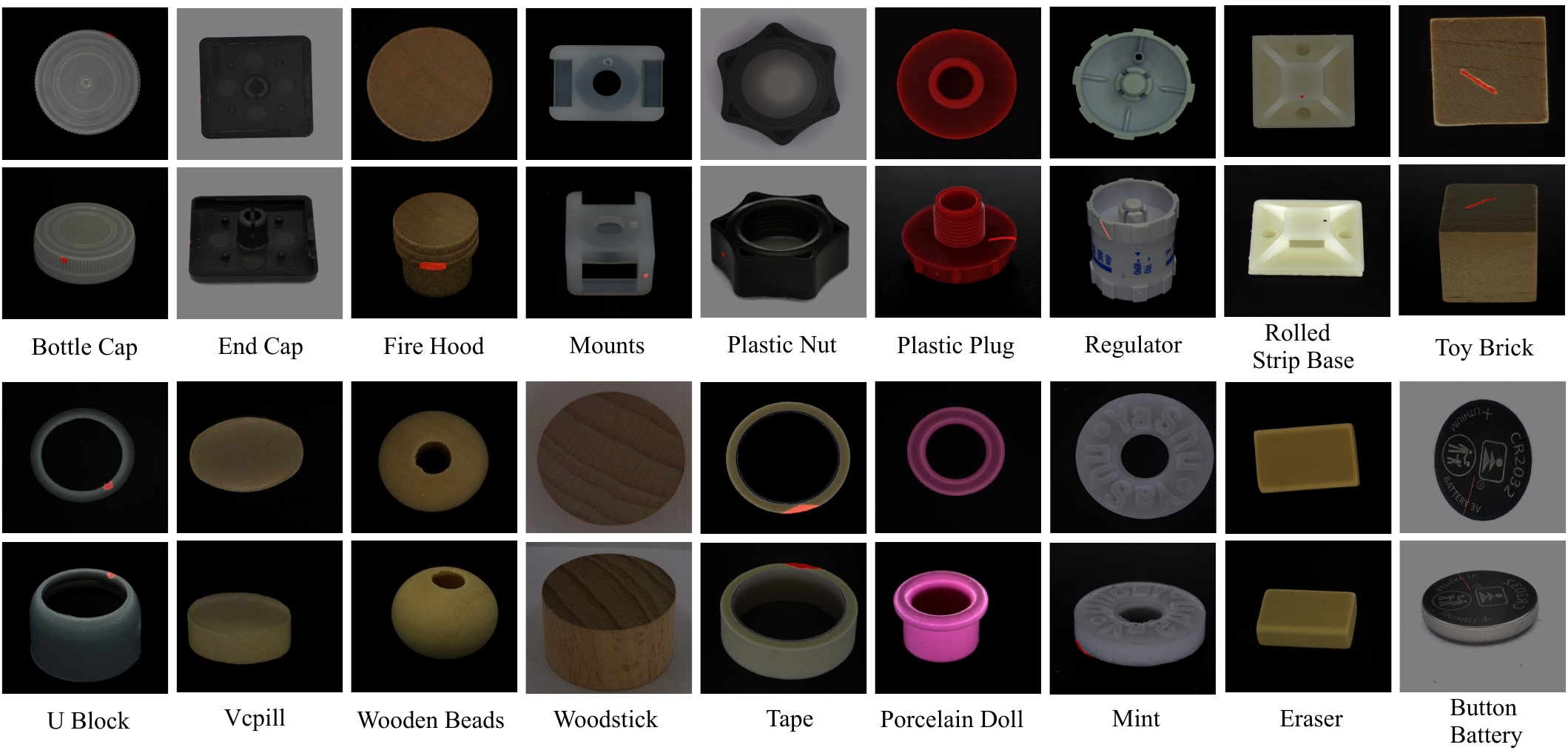}
  }
  \caption{\textbf{Real-IAD Visualisation}. (a) The upper six rows show 12 complex objects and all 5 views of each object. (b) The bottom 4 rows show the remaining 18 simple objects and each object shows two views. All anomaly regions are marked in red. Best viewed in zoom and color.}
  \label{fig:vis_realiad}
\end{figure*}

\twocolumn[{
\renewcommand\twocolumn[2][]{#1}%
In this supplementary material, we first give all 30 object visualizations including 12 complex products (mixed material and surfaces with complex geometries) and 18 simple products (single material, e.g., metal, plastic, 
wood, ceramics and etc., and simple and smooth surfaces) in Real-IAD in Fig~\ref{fig:vis_realiad}. Then, some representative anomaly images covering all 8 defects, i.e., pit, deformation, abrasion, scratch, damage, missing, foreign objects, and contamination, are shown in Fig~\ref{fig:defect_realiad}. In our main paper, we only report averaged results for all methods, here complete experimental results of all categories with both UIAD and FUIAD settings are reported in Tabs.~\ref{tab:sv-uiad}, \ref{tab:mv-uiad}, \ref{tab:mv-fuiad-0}, \ref{tab:mv-fuiad-1}, \ref{tab:mv-fuiad-2} and \ref{tab:mv-fuiad-4}.

\vspace{20pt}

\centering
{
    \includegraphics[width=1.0\linewidth]{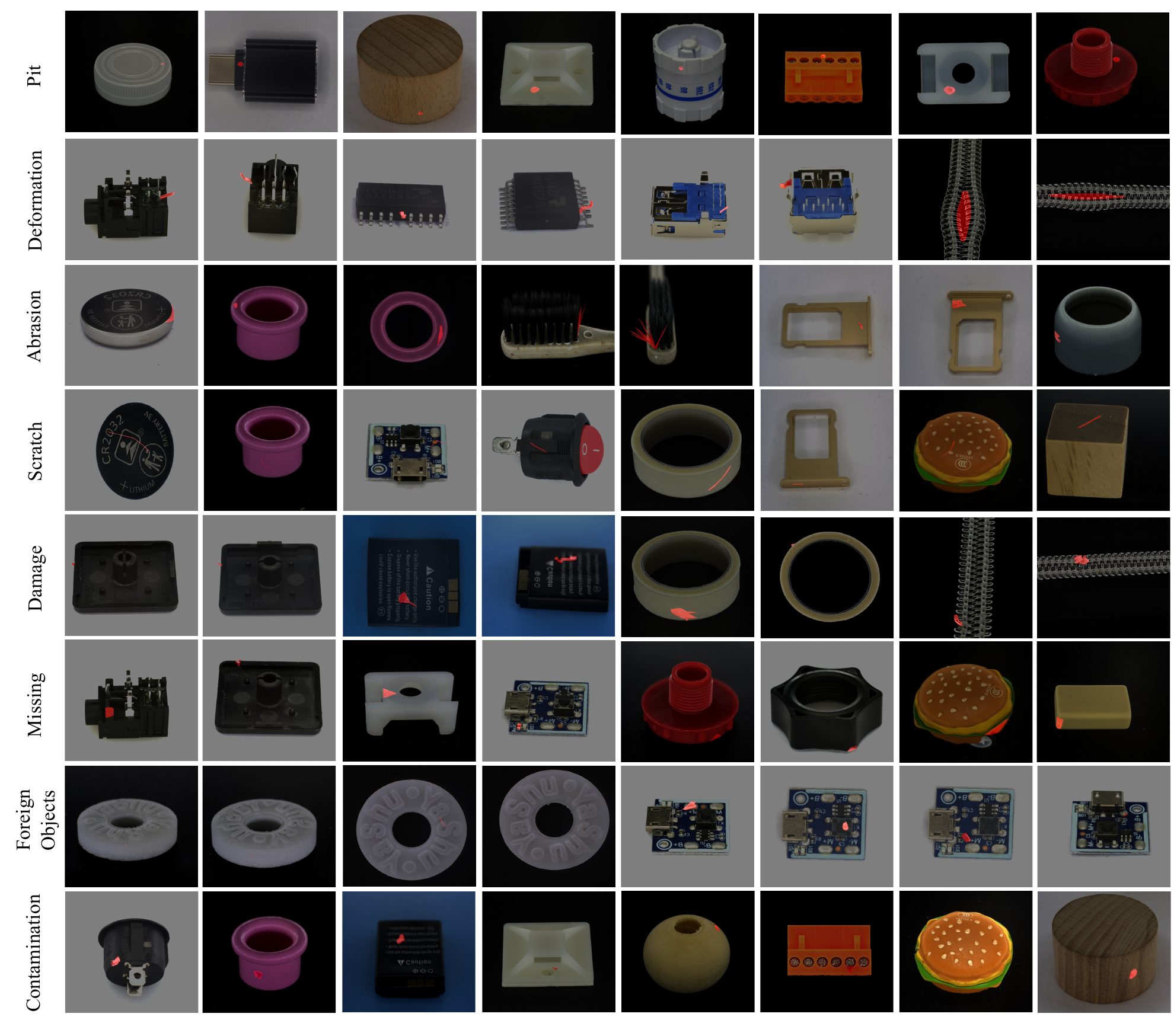}
}
\captionof{figure}{\textbf{Anomalies Visualization of Real-IAD}. Each row represents an anomaly class from different products and diverse locations. From top to bottom, they are pit, deformation, abrasion, scratch, damage, missing, foreign objects, and contamination, respectively.}\label{fig:defect_realiad}
}]

\begin{table*}[htb]
    \centering
    \caption{UIAD performance~(I-AUROC / P-AUROC) comparisons with state-of-the-art anomaly detection methods on single-view Real-IAD. 
    }
    \label{tab:sv-uiad}
    \renewcommand{\arraystretch}{1.2}
    \setlength\tabcolsep{6.0pt}
    \resizebox{1.0\linewidth}{!}{
        \begin{tabular}{ll | cc cc | cc | cc}
            \hline
            \multicolumn{2}{c|}{\multirow{2}{*}{Category}} & \multicolumn{4}{c|}{Embedding-based} & \multicolumn{2}{c|}{Data-Aug-based} & \multicolumn{2}{c}{Reconstruction-based} \\ 
            \cline{3-6} \cline{7-8} \cline{9-10}
            & & \textbf{PaDim\cite{padim}} &\textbf{CFlow}\cite{cflow} &\textbf{PatchCore\cite{patchcore}}   &\textbf{SoftPatch\cite{softpatch}} & \textbf{SimpleNet\cite{simplenet}} & \textbf{DeSTSeg\cite{destseg}} & \textbf{RD\cite{rd}} & \textbf{UniAD\cite{uniad}} \\
            \hline
            \multirow{30}{*}{\rotatebox{90}{Top-view UAD}} & Audiojack & 81.6 / 89.6 & 77.0 / 85.4 & 84.0 / 88.7 & 83.8 / 92.2 & 88.1 / 90.7 & 88.0 / 94.2 & 83.5 / 91.2 & 82.2 / 85.7 \\
& PCB & 88.6 / 79.0 & 83.0 / 85.2 & 93.9 / 94.7 & 91.3 / 95.0 & 91.9 / 91.5 & 88.5 / 92.8 & 89.6 / 93.8 & 82.5 / 78.2 \\
& Phone Battery & 87.6 / 85.4 & 85.3 / 89.7 & 91.9 / 91.2 & 90.4 / 79.3 & 92.7 / 75.9 & 92.6 / 94.5 & 90.4 / 94.7 & 84.5 / 85.1 \\
& SIM Card Set & 94.8 / 95.2 & 97.1 / 94.7 & 98.5 / 97.2 & 96.2 / 88.7 & 98.6 / 86.4 & 95.0 / 94.3 & 91.9 / 97.3 & 94.6 / 91.5 \\
& Switch & 89.1 / 92.8 & 86.4 / 85.6 & 91.4 / 93.7 & 95.7 / 96.4 & 97.3 / 95.5 & 95.6 / 97.4 & 91.8 / 93.7 & 85.0 / 90.8 \\
& Terminal Block & 93.3 / 98.4 & 89.7 / 93.5 & 97.9 / 98.5 & 97.8 / 98.8 & 98.8 / 97.9 & 95.9 / 96.9 & 97.4 / 98.8 & 90.7 / 96.8 \\
& Toothbrush & 87.9 / 92.2 & 87.9 / 90.1 & 90.0 / 95.2 & 88.5 / 86.9 & 90.6 / 82.1 & 92.3 / 88.1 & 85.6 / 95.3 & 89.4 / 88.3 \\
& Toy & 94.1 / 83.0 & 90.5 / 77.9 & 89.2 / 90.1 & 91.4 / 92.4 & 91.1 / 87.3 & 85.2 / 83.8 & 86.6 / 93.9 & 74.5 / 79.8 \\
& Transistor1 & 90.2 / 91.4 & 86.9 / 90.2 & 95.9 / 97.0 & 95.0 / 97.2 & 97.6 / 96.2 & 96.6 / 92.5 & 93.5 / 96.5 & 90.9 / 89.7 \\
& USB & 84.5 / 92.0 & 91.2 / 92.0 & 95.0 / 97.1 & 96.2 / 98.1 & 97.2 / 97.2 & 90.7 / 95.9 & 88.5 / 98.0 & 88.1 / 90.7 \\
& USB Adaptor & 86.7 / 87.2 & 75.8 / 87.3 & 82.7 / 89.8 & 86.6 / 88.9 & 86.5 / 85.7 & 84.0 / 83.9 & 78.8 / 91.3 & 76.4 / 71.3 \\
& Zipper & 94.4 / 90.2 & 95.6 / 93.4 & 98.8 / 96.1 & 98.1 / 95.4 & 99.6 / 94.5 & 99.9 / 95.8 & 99.6 / 98.1 & 95.4 / 94.0 \\
& Bottle Cap & 89.2 / 94.7 & 88.9 / 91.7 & 92.2 / 94.9 & 95.2 / 96.2 & 95.0 / 95.4 & 87.3 / 93.2 & 90.5 / 92.6 & 87.4 / 89.5 \\
& End Cap & 73.5 / 86.2 & 65.9 / 80.3 & 76.8 / 89.6 & 76.2 / 91.5 & 81.2 / 85.4 & 78.4 / 92.7 & 69.7 / 91.2 & 73.5 / 82.5 \\
& Fire Hood & 92.2 / 94.9 & 92.0 / 95.5 & 93.4 / 94.0 & 93.4 / 93.8 & 94.9 / 92.8 & 85.5 / 93.3 & 92.5 / 96.9 & 93.1 / 93.5 \\
& Mounts & 89.0 / 96.9 & 87.8 / 94.6 & 92.8 / 97.8 & 91.2 / 98.0 & 95.1 / 97.1 & 95.3 / 97.5 & 91.1 / 98.1 & 84.8 / 97.1 \\
& Plastic Nut & 76.1 / 81.3 & 77.9 / 81.8 & 84.7 / 82.9 & 81.8 / 95.5 & 85.8 / 88.4 & 85.1 / 90.0 & 82.6 / 93.2 & 71.3 / 82.8 \\
& Plastic Plug & 93.0 / 96.7 & 92.0 / 96.0 & 95.3 / 96.9 & 94.5 / 97.6 & 94.4 / 92.3 & 91.9 / 97.0 & 90.9 / 97.2 & 81.3 / 92.7 \\
& Regulator & 83.1 / 98.5 & 71.1 / 82.0 & 96.7 / 99.6 & 89.5 / 99.2 & 97.9 / 98.3 & 98.1 / 97.9 & 87.8 / 98.9 & 53.8 / 74.2 \\
& Rolled Strip Base & 99.1 / 95.4 & 98.3 / 98.1 & 99.3 / 98.5 & 98.9 / 98.9 & 99.8 / 97.1 & 99.2 / 88.6 & 94.9 / 96.1 & 97.7 / 97.6 \\
& Toy Brick & 78.2 / 83.5 & 83.4 / 82.7 & 84.0 / 84.8 & 82.6 / 85.9 & 84.3 / 80.6 & 80.0 / 81.6 & 82.1 / 90.1 & 84.6 / 86.7 \\
& U Block & 89.3 / 97.2 & 85.0 / 95.9 & 93.4 / 98.7 & 91.6 / 97.9 & 93.4 / 95.1 & 90.0 / 96.7 & 88.5 / 98.1 & 80.4 / 88.4 \\
& Vcpill & 84.2 / 91.3 & 91.7 / 90.0 & 92.8 / 91.6 & 89.1 / 83.2 & 92.0 / 76.9 & 91.1 / 74.8 & 90.6 / 94.9 & 81.9 / 89.3 \\
& Wooden Beads & 86.3 / 86.1 & 84.0 / 84.4 & 84.4 / 84.4 & 85.4 / 85.3 & 85.2 / 78.7 & 86.4 / 83.6 & 78.8 / 85.6 & 81.3 / 76.7 \\
& Woodstick & 89.9 / 83.2 & 86.0 / 93.8 & 82.8 / 77.2 & 80.6 / 75.8 & 86.2 / 87.4 & 84.9 / 93.4 & 89.4 / 94.9 & 89.0 / 91.1 \\
& Tape & 96.4 / 98.2 & 95.8 / 97.7 & 98.7 / 99.0 & 98.9 / 99.2 & 98.7 / 97.9 & 98.9 / 98.4 & 98.2 / 98.8 & 97.7 / 97.6 \\
& Porcelain Doll & 79.3 / 88.9 & 84.3 / 93.4 & 88.6 / 96.0 & 85.8 / 94.7 & 87.0 / 88.2 & 83.3 / 75.7 & 87.5 / 95.4 & 66.9 / 84.7 \\
& Mint & 57.6 / 66.6 & 57.8 / 62.0 & 69.5 / 72.4 & 63.7 / 70.1 & 69.4 / 61.6 & 64.1 / 66.1 & 63.6 / 78.3 & 56.7 / 44.0 \\
& Eraser & 90.6 / 94.3 & 90.2 / 91.1 & 92.9 / 95.5 & 91.8 / 94.5 & 93.2 / 93.3 & 89.6 / 90.0 & 87.5 / 95.4 & 90.4 / 91.0 \\
& Button Battery & 78.3 / 88.2 & 72.5 / 80.9 & 85.5 / 91.2 & 86.6 / 86.7 & 87.7 / 81.1 & 81.1 / 90.0 & 83.6 / 94.1 & 74.1 / 78.2 \\
\cline{2-10}
& Average All & 86.6 / 90.0 & 85.0 / 88.6 & 90.4 / 92.5 & 89.6 / 91.8 & 91.7 / 88.9 & 89.2 / 90.4 & 87.6 / 94.4 & 82.7 / 86.0 \\

            \hline
        \end{tabular}
}
\end{table*}

\begin{table*}[htb]
    \centering
    \caption{UIAD performance~(S-AUROC / I-AUROC / P-AUPRO) comparisons with state-of-the-art anomaly detection methods on multi-view Real-IAD. 
    }
    \label{tab:mv-uiad}
    \renewcommand{\arraystretch}{1.2}
    \setlength\tabcolsep{6.0pt}
    \resizebox{1.0\linewidth}{!}{
        \begin{tabular}{ll | cc cc | cc | cc}
            \hline
            \multicolumn{2}{c|}{\multirow{2}{*}{Category}} & \multicolumn{4}{c|}{Embedding-based} & \multicolumn{2}{c|}{Data-Aug-based} & \multicolumn{2}{c}{Reconstruction-based} \\ 
            \cline{3-6} \cline{7-8} \cline{9-10}
            & & \textbf{PaDim\cite{padim}} &\textbf{CFlow}\cite{cflow} &\textbf{PatchCore\cite{patchcore}}   &\textbf{SoftPatch\cite{softpatch}} & \textbf{SimpleNet\cite{simplenet}} & \textbf{DeSTSeg\cite{destseg}} & \textbf{RD\cite{rd}} & \textbf{UniAD\cite{uniad}} \\
            \hline
            \multirow{30}{*}{\rotatebox{90}{Multi-view UAD}} & Audiojack & 92.2 / 66.6 / 79.5 & 82.0 / 74.3 / 83.4 & 89.3 / 86.3 / 87.6 & 91.0 / 88.5 / 92.5 & 91.2 / 87.4 / 89.3 & 95.3 / 81.8 / 58.8 & 81.9 / 82.4 / 87.7 & 91.2 / 82.8 / 84.0 \\
& PCB & 88.4 / 75.0 / 65.5 & 83.1 / 77.3 / 87.3 & 93.0 / 92.4 / 93.5 & 90.3 / 90.8 / 93.9 & 90.7 / 87.1 / 88.0 & 83.6 / 79.6 / 86.4 & 79.3 / 89.5 / 94.2 & 83.2 / 80.5 / 80.4 \\
& Phone Battery & 91.7 / 81.9 / 83.3 & 91.2 / 84.4 / 92.0 & 95.1 / 91.6 / 91.6 & 91.5 / 90.2 / 83.3 & 94.7 / 88.9 / 72.8 & 98.2 / 86.2 / 77.9 & 89.4 / 90.8 / 95.8 & 93.6 / 83.4 / 87.3 \\
& SIM Card Set & 94.2 / 91.8 / 84.1 & 95.6 / 94.4 / 92.2 & 99.3 / 97.1 / 87.7 & 98.4 / 95.8 / 75.9 & 99.2 / 96.1 / 72.8 & 98.3 / 91.8 / 91.8 & 89.9 / 93.6 / 92.7 & 94.0 / 91.1 / 83.4 \\
& Switch & 82.1 / 80.8 / 88.8 & 92.9 / 83.9 / 87.5 & 94.6 / 89.4 / 93.6 & 97.8 / 94.7 / 96.2 & 98.8 / 94.8 / 93.8 & 96.6 / 92.5 / 94.3 & 87.3 / 87.2 / 93.7 & 95.3 / 85.7 / 90.7 \\
& Terminal Block & 96.9 / 86.7 / 95.1 & 92.2 / 85.1 / 96.4 & 97.5 / 94.1 / 97.1 & 98.2 / 96.2 / 97.8 & 97.7 / 93.4 / 95.0 & 96.1 / 91.4 / 90.2 & 89.8 / 94.5 / 98.6 & 93.8 / 85.9 / 93.1 \\
& Toothbrush & 91.7 / 74.4 / 75.2 & 91.9 / 79.9 / 85.8 & 94.7 / 88.0 / 91.9 & 92.9 / 89.3 / 87.4 & 95.3 / 86.1 / 76.6 & 97.9 / 87.5 / 87.0 & 86.7 / 83.4 / 90.9 & 95.0 / 82.7 / 86.6 \\
& Toy & 91.4 / 82.5 / 72.4 & 78.8 / 67.8 / 81.2 & 92.8 / 87.3 / 91.2 & 91.3 / 86.7 / 91.9 & 92.9 / 82.2 / 77.5 & 96.5 / 82.1 / 78.5 & 75.0 / 81.5 / 92.4 & 77.2 / 67.3 / 74.4 \\
& Transistor1 & 90.3 / 86.5 / 88.3 & 96.6 / 92.8 / 94.8 & 99.8 / 97.8 / 97.5 & 99.3 / 97.2 / 97.6 & 99.7 / 97.3 / 95.5 & 99.0 / 95.6 / 96.4 & 94.7 / 95.8 / 97.6 & 99.3 / 94.1 / 92.9 \\
& USB & 77.0 / 69.7 / 76.9 & 86.1 / 78.1 / 83.7 & 93.9 / 90.7 / 95.1 & 93.8 / 91.2 / 96.8 & 93.9 / 90.0 / 93.5 & 93.3 / 90.3 / 89.1 & 89.4 / 92.1 / 96.1 & 83.1 / 80.9 / 85.2 \\
& USB Adaptor & 93.2 / 81.9 / 89.9 & 86.8 / 75.2 / 88.8 & 90.6 / 83.8 / 91.4 & 91.9 / 86.0 / 91.8 & 93.0 / 82.7 / 79.4 & 93.6 / 73.1 / 55.3 & 65.3 / 71.9 / 89.5 & 85.1 / 77.3 / 79.9 \\
& Zipper & 99.3 / 90.4 / 78.5 & 97.6 / 93.8 / 94.2 & 100.0 / 98.1 / 90.8 & 99.6 / 97.9 / 90.9 & 99.7 / 98.2 / 92.4 & 99.7 / 98.4 / 99.0 & 96.1 / 97.8 / 98.0 & 98.8 / 98.2 / 94.7 \\
& Bottle Cap & 98.1 / 87.7 / 91.9 & 98.8 / 91.3 / 95.4 & 99.4 / 94.3 / 93.0 & 99.1 / 95.9 / 96.8 & 99.4 / 91.6 / 88.2 & 92.4 / 87.1 / 83.4 & 93.7 / 89.2 / 97.0 & 97.3 / 89.8 / 93.8 \\
& End Cap & 76.1 / 73.7 / 85.7 & 75.1 / 66.4 / 87.2 & 91.9 / 84.1 / 92.2 & 92.3 / 85.8 / 93.6 & 94.2 / 83.4 / 82.9 & 82.3 / 80.3 / 81.6 & 68.1 / 79.0 / 93.4 & 89.4 / 80.4 / 85.6 \\
& Fire Hood & 96.9 / 77.3 / 72.7 & 87.0 / 80.5 / 93.5 & 89.3 / 84.1 / 88.6 & 87.8 / 84.3 / 89.2 & 95.6 / 81.8 / 82.7 & 96.9 / 78.9 / 62.7 & 81.4 / 84.3 / 91.7 & 83.0 / 79.5 / 84.5 \\
& Mounts & 98.4 / 82.5 / 89.1 & 98.2 / 85.3 / 95.6 & 99.7 / 88.0 / 91.3 & 99.3 / 85.9 / 89.6 & 99.4 / 88.2 / 89.2 & 99.1 / 85.1 / 68.6 & 92.5 / 85.7 / 94.6 & 97.0 / 87.2 / 94.3 \\
& Plastic Nut & 98.2 / 73.8 / 87.4 & 88.6 / 79.8 / 94.1 & 97.8 / 90.8 / 95.4 & 95.7 / 89.3 / 96.2 & 95.7 / 89.3 / 90.6 & 94.4 / 90.2 / 81.3 & 72.8 / 85.0 / 96.7 & 87.1 / 79.3 / 87.2 \\
& Plastic Plug & 87.4 / 80.0 / 86.7 & 90.0 / 83.9 / 93.3 & 95.7 / 89.7 / 92.2 & 92.5 / 88.7 / 92.9 & 94.4 / 87.1 / 84.4 & 95.6 / 86.0 / 74.4 & 89.3 / 90.5 / 96.3 & 78.0 / 78.2 / 87.7 \\
& Regulator & 96.5 / 76.5 / 85.1 & 85.1 / 62.9 / 85.0 & 86.0 / 81.9 / 95.7 & 82.9 / 82.1 / 95.1 & 92.0 / 82.2 / 88.7 & 93.0 / 89.8 / 79.9 & 92.5 / 87.3 / 97.0 & 55.5 / 51.8 / 70.3 \\
& Rolled Strip Base & 98.6 / 97.4 / 96.1 & 98.8 / 97.1 / 98.4 & 99.7 / 98.9 / 98.3 & 99.7 / 99.1 / 99.0 & 99.6 / 99.4 / 95.7 & 98.9 / 98.3 / 98.5 & 80.3 / 94.3 / 98.5 & 99.3 / 98.6 / 97.7 \\
& Toy Brick & 84.3 / 71.0 / 76.9 & 82.9 / 80.7 / 88.7 & 82.6 / 81.4 / 83.6 & 78.2 / 79.8 / 85.7 & 85.7 / 80.8 / 75.7 & 87.0 / 76.8 / 70.0 & 72.5 / 73.9 / 88.7 & 78.3 / 77.9 / 82.3 \\
& U Block & 98.3 / 84.1 / 89.8 & 96.7 / 90.5 / 96.7 & 98.8 / 91.9 / 96.8 & 98.3 / 91.6 / 96.5 & 98.5 / 90.7 / 91.7 & 98.5 / 90.1 / 81.4 & 86.9 / 91.7 / 97.1 & 96.3 / 88.8 / 94.0 \\
& Vcpill & 94.7 / 67.4 / 80.1 & 87.8 / 85.8 / 93.5 & 96.5 / 91.4 / 93.5 & 93.7 / 89.6 / 88.7 & 97.5 / 91.1 / 85.2 & 96.4 / 90.0 / 89.2 & 87.2 / 90.2 / 94.3 & 89.4 / 88.6 / 94.2 \\
& Wooden Beads & 91.1 / 82.4 / 78.3 & 89.3 / 87.4 / 89.4 & 91.4 / 88.6 / 87.3 & 90.9 / 89.2 / 87.9 & 92.9 / 85.7 / 81.6 & 91.9 / 86.2 / 82.7 & 85.0 / 87.4 / 89.2 & 82.5 / 80.7 / 83.0 \\
& Woodstick & 81.8 / 79.9 / 78.8 & 83.9 / 78.9 / 89.0 & 74.5 / 77.2 / 69.1 & 73.9 / 76.0 / 70.8 & 81.5 / 77.9 / 71.7 & 90.2 / 89.0 / 79.2 & 71.9 / 84.2 / 91.6 & 76.0 / 78.9 / 77.2 \\
& Tape & 99.8 / 93.7 / 96.4 & 98.5 / 96.0 / 98.8 & 99.9 / 98.2 / 97.9 & 99.7 / 97.9 / 97.9 & 100.0 / 96.9 / 95.4 & 99.1 / 96.1 / 92.8 & 89.5 / 93.1 / 98.7 & 99.1 / 97.2 / 97.6 \\
& Porcelain Doll & 93.8 / 74.3 / 75.2 & 95.1 / 76.0 / 95.4 & 96.1 / 88.2 / 94.3 & 94.7 / 86.1 / 93.4 & 96.2 / 86.1 / 86.0 & 94.6 / 83.7 / 68.7 & 89.6 / 87.8 / 95.1 & 92.8 / 84.1 / 91.7 \\
& Mint & 69.1 / 66.9 / 61.4 & 79.1 / 70.7 / 75.1 & 85.7 / 76.2 / 81.4 & 82.1 / 74.5 / 80.8 & 86.8 / 77.2 / 66.0 & 77.7 / 70.5 / 71.2 & 67.7 / 71.6 / 79.9 & 73.0 / 67.6 / 60.2 \\
& Eraser & 96.5 / 86.7 / 90.7 & 87.2 / 88.1 / 97.0 & 95.6 / 93.4 / 96.1 & 96.1 / 93.5 / 95.9 & 94.7 / 91.2 / 90.0 & 91.9 / 88.2 / 83.3 & 82.9 / 89.2 / 96.8 & 91.2 / 89.6 / 93.8 \\
& Button Battery & 88.7 / 84.8 / 54.1 & 96.3 / 85.2 / 82.3 & 90.6 / 87.3 / 88.2 & 91.9 / 88.5 / 78.9 & 95.8 / 88.8 / 64.0 & 93.3 / 91.2 / 90.4 & 83.3 / 87.0 / 91.1 & 87.5 / 79.0 / 75.8 \\
\cline{2-10}
& Average All & 91.2 / 80.3 / 81.8 & 89.8 / 82.5 / 90.5 & 93.7 / 89.4 / 91.5 & 92.8 / 89.4 / 90.8 & 94.9 / 88.5 / 84.6 & 94.0 / 86.9 / 81.5 & 83.7 / 87.1 / 93.8 & 88.1 / 82.9 / 86.1 \\
            \hline
        \end{tabular}
}
\end{table*}

\begin{table*}[thb]
    \centering
    \caption{FUIAD performance~(S-AUROC / I-AUROC / P-AUPRO) comparisons with state-of-the-art anomaly detection methods on multi-view Real-IAD with a noisy ratio of 0. 
    }
    \label{tab:mv-fuiad-0}
    \renewcommand{\arraystretch}{1.2}
    \setlength\tabcolsep{6.0pt}
    \resizebox{1.0\linewidth}{!}{
        \begin{tabular}{ll | cc cc | cc | cc}
            \hline
            \multicolumn{2}{c|}{\multirow{2}{*}{Category}} & \multicolumn{4}{c|}{Embedding-based} & \multicolumn{2}{c|}{Data-Aug-based} & \multicolumn{2}{c}{Reconstruction-based} \\ 
            \cline{3-6} \cline{7-8} \cline{9-10}
            & & \textbf{PaDim\cite{padim}} &\textbf{CFlow}\cite{cflow} &\textbf{PatchCore\cite{patchcore}}   &\textbf{SoftPatch\cite{softpatch}} & \textbf{SimpleNet\cite{simplenet}} & \textbf{DeSTSeg\cite{destseg}} & \textbf{RD\cite{rd}} & \textbf{UniAD\cite{uniad}} \\
            \hline
            \multirow{30}{*}{\rotatebox{90}{FUAD-NR-0.0}} & Audiojack & 96.0 / 74.1 / 83.5 & 80.4 / 78.0 / 87.1 & 93.0 / 89.0 / 89.7 & 93.0 / 90.4 / 92.9 & 94.7 / 90.4 / 89.7 & 96.7 / 85.4 / 88.6 & 84.6 / 86.6 / 90.9 & 95.0 / 87.1 / 86.6 \\
& PCB & 92.0 / 79.8 / 70.0 & 81.7 / 78.3 / 87.0 & 94.4 / 94.1 / 94.1 & 94.4 / 94.2 / 95.7 & 93.9 / 89.4 / 90.3 & 94.0 / 92.9 / 91.9 & 94.3 / 94.0 / 94.2 & 87.8 / 85.8 / 84.1 \\
& Phone Battery & 92.0 / 86.0 / 87.3 & 97.4 / 86.3 / 94.0 & 96.4 / 92.7 / 94.2 & 95.3 / 92.7 / 86.2 & 98.3 / 90.1 / 70.3 & 98.4 / 89.7 / 94.0 & 80.4 / 91.4 / 97.7 & 92.6 / 87.6 / 91.5 \\
& SIM Card Set & 95.0 / 95.2 / 89.1 & 91.7 / 91.9 / 93.7 & 98.8 / 98.7 / 91.7 & 98.0 / 97.9 / 79.8 & 98.4 / 97.6 / 72.5 & 96.1 / 93.9 / 90.5 & 85.2 / 95.5 / 94.4 & 94.4 / 94.1 / 85.9 \\
& Switch & 83.0 / 81.9 / 89.5 & 92.5 / 87.6 / 89.4 & 94.7 / 89.6 / 93.8 & 98.4 / 96.1 / 96.7 & 98.8 / 94.8 / 94.2 & 97.2 / 94.5 / 94.2 & 91.7 / 91.8 / 94.7 & 95.6 / 88.3 / 92.3 \\
& Terminal Block & 95.6 / 91.6 / 95.8 & 92.6 / 89.0 / 95.9 & 97.1 / 95.2 / 97.3 & 97.6 / 96.4 / 97.8 & 96.5 / 94.1 / 95.1 & 97.3 / 94.9 / 96.9 & 85.0 / 94.6 / 98.6 & 96.0 / 89.6 / 94.1 \\
& Toothbrush & 93.6 / 77.0 / 77.1 & 96.0 / 82.9 / 86.1 & 95.1 / 88.7 / 92.6 & 92.8 / 90.0 / 88.9 & 94.6 / 87.1 / 76.1 & 99.0 / 89.6 / 85.2 & 77.8 / 83.2 / 92.1 & 94.9 / 83.7 / 88.2 \\
& Toy & 92.7 / 88.9 / 75.8 & 73.0 / 65.2 / 78.4 & 95.5 / 90.5 / 92.3 & 94.2 / 90.0 / 92.6 & 94.8 / 85.0 / 73.6 & 98.0 / 88.4 / 77.5 & 84.9 / 86.5 / 93.2 & 78.7 / 69.0 / 77.4 \\
& Transistor1 & 95.6 / 90.7 / 90.8 & 99.6 / 95.8 / 95.8 & 99.9 / 98.4 / 98.0 & 99.8 / 97.9 / 98.0 & 100.0 / 97.7 / 95.7 & 99.3 / 95.5 / 90.0 & 99.9 / 98.2 / 97.8 & 99.6 / 95.5 / 94.2 \\
& USB & 80.5 / 74.4 / 82.5 & 79.4 / 78.2 / 87.9 & 95.4 / 93.9 / 96.4 & 94.2 / 94.0 / 98.1 & 94.3 / 94.1 / 94.6 & 93.9 / 92.8 / 93.9 & 85.7 / 93.2 / 97.4 & 84.2 / 84.4 / 88.4 \\
& USB Adaptor & 92.7 / 85.8 / 91.7 & 87.4 / 76.0 / 89.3 & 90.1 / 84.2 / 92.0 & 92.6 / 88.6 / 92.3 & 90.9 / 84.5 / 79.5 & 93.6 / 82.9 / 85.7 & 75.1 / 75.7 / 90.6 & 86.8 / 81.3 / 84.1 \\
& Zipper & 99.3 / 92.1 / 80.5 & 97.7 / 96.4 / 93.7 & 100.0 / 97.9 / 91.0 & 99.8 / 97.5 / 91.2 & 99.8 / 98.0 / 91.2 & 99.7 / 99.2 / 95.0 & 90.7 / 97.5 / 98.4 & 99.0 / 98.3 / 96.2 \\
& Bottle Cap & 98.5 / 93.2 / 95.5 & 98.5 / 93.6 / 97.0 & 99.3 / 96.8 / 94.5 & 99.4 / 97.6 / 98.3 & 98.2 / 94.6 / 93.4 & 97.2 / 91.4 / 92.1 & 92.0 / 94.9 / 98.0 & 99.2 / 94.4 / 96.5 \\
& End Cap & 79.1 / 74.6 / 85.7 & 79.3 / 68.1 / 85.3 & 94.4 / 85.8 / 92.4 & 95.1 / 87.8 / 93.9 & 95.9 / 84.3 / 76.2 & 88.2 / 83.7 / 81.4 & 66.1 / 76.1 / 93.7 & 91.1 / 80.7 / 84.3 \\
& Fire Hood & 97.8 / 84.2 / 76.4 & 87.8 / 84.3 / 93.6 & 88.3 / 85.4 / 90.0 & 87.9 / 86.2 / 91.1 & 93.2 / 82.0 / 80.4 & 92.0 / 85.5 / 88.1 & 68.8 / 82.1 / 93.3 & 81.4 / 79.9 / 86.4 \\
& Mounts & 99.4 / 87.2 / 88.2 & 98.6 / 88.8 / 95.7 & 100.0 / 91.1 / 90.1 & 100.0 / 90.0 / 87.9 & 99.5 / 89.6 / 89.4 & 99.7 / 87.3 / 87.4 & 86.7 / 88.8 / 94.4 & 97.8 / 89.5 / 93.5 \\
& Plastic Nut & 98.8 / 79.3 / 89.4 & 89.7 / 82.1 / 92.9 & 98.9 / 94.7 / 96.1 & 96.6 / 93.6 / 96.9 & 96.3 / 93.0 / 90.2 & 93.7 / 92.7 / 96.4 & 93.2 / 92.5 / 96.9 & 89.0 / 82.7 / 89.4 \\
& Plastic Plug & 85.7 / 85.9 / 87.5 & 82.7 / 81.4 / 94.1 & 93.4 / 92.9 / 94.0 & 91.1 / 91.6 / 94.7 & 95.4 / 90.6 / 84.1 & 93.4 / 90.7 / 92.9 & 91.7 / 93.7 / 96.3 & 83.4 / 83.5 / 87.9 \\
& Regulator & 97.8 / 83.1 / 86.7 & 81.0 / 66.3 / 83.0 & 90.7 / 86.4 / 96.9 & 89.8 / 86.3 / 96.5 & 94.3 / 81.8 / 85.0 & 99.0 / 92.6 / 95.2 & 93.4 / 89.4 / 98.3 & 64.7 / 61.0 / 74.3 \\
& Rolled Strip Base & 98.9 / 98.0 / 97.3 & 97.9 / 95.9 / 98.9 & 99.4 / 99.0 / 98.9 & 99.6 / 99.3 / 99.2 & 99.6 / 99.3 / 96.8 & 98.8 / 98.7 / 98.0 & 98.2 / 97.0 / 98.7 & 99.1 / 98.5 / 98.2 \\
& Toy Brick & 88.3 / 73.2 / 77.0 & 81.6 / 79.3 / 88.1 & 87.1 / 83.5 / 84.8 & 82.9 / 81.1 / 87.0 & 82.6 / 74.9 / 66.4 & 73.9 / 69.4 / 60.1 & 78.3 / 78.7 / 89.0 & 81.2 / 77.6 / 82.5 \\
& U Block & 98.0 / 88.2 / 92.1 & 95.6 / 91.8 / 95.5 & 98.2 / 93.7 / 97.0 & 97.4 / 93.2 / 96.8 & 97.7 / 93.1 / 92.5 & 98.7 / 89.3 / 93.6 & 85.8 / 92.6 / 98.1 & 95.1 / 91.2 / 94.0 \\
& Vcpill & 93.9 / 73.4 / 82.6 & 87.7 / 88.1 / 94.0 & 96.0 / 92.4 / 94.6 & 94.2 / 90.5 / 90.5 & 97.8 / 91.6 / 85.9 & 97.0 / 89.1 / 81.5 & 96.0 / 89.9 / 95.7 & 86.5 / 87.7 / 93.9 \\
& Wooden Beads & 92.7 / 89.8 / 84.8 & 89.9 / 90.9 / 90.8 & 89.4 / 89.7 / 89.2 & 89.2 / 90.1 / 89.9 & 90.0 / 87.1 / 82.7 & 88.7 / 88.2 / 88.2 & 83.6 / 88.9 / 91.6 & 81.9 / 81.6 / 85.6 \\
& Woodstick & 83.7 / 82.9 / 79.8 & 80.2 / 80.8 / 88.7 & 75.5 / 81.1 / 71.0 & 74.8 / 80.6 / 73.3 & 79.7 / 82.1 / 71.6 & 89.1 / 88.9 / 90.3 & 61.9 / 84.4 / 92.8 & 79.1 / 82.9 / 78.4 \\
& Tape & 99.8 / 95.7 / 96.9 & 96.8 / 94.7 / 99.3 & 99.7 / 98.8 / 98.0 & 99.6 / 98.5 / 98.0 & 99.9 / 97.6 / 96.3 & 99.7 / 96.4 / 98.3 & 97.2 / 98.2 / 99.3 & 99.6 / 98.2 / 98.6 \\
& Porcelain Doll & 96.3 / 81.0 / 80.2 & 91.0 / 79.9 / 95.4 & 96.7 / 91.0 / 96.1 & 95.8 / 88.2 / 95.4 & 98.3 / 90.0 / 88.4 & 95.9 / 87.1 / 90.7 & 95.1 / 89.4 / 96.9 & 94.0 / 86.8 / 93.4 \\
& Mint & 76.7 / 71.2 / 64.0 & 79.7 / 71.4 / 71.6 & 90.7 / 80.2 / 83.2 & 88.2 / 77.5 / 83.1 & 91.0 / 81.7 / 64.6 & 85.4 / 78.0 / 64.0 & 70.6 / 74.5 / 86.4 & 79.2 / 70.0 / 57.8 \\
& Eraser & 96.8 / 90.7 / 94.2 & 86.0 / 89.2 / 98.2 & 95.4 / 94.9 / 98.1 & 95.0 / 95.3 / 97.7 & 93.5 / 92.0 / 86.4 & 94.5 / 91.2 / 93.4 & 89.4 / 92.7 / 97.8 & 90.4 / 91.6 / 94.8 \\
& Button Battery & 90.6 / 89.2 / 59.1 & 90.2 / 84.0 / 78.6 & 87.9 / 87.5 / 90.0 & 90.3 / 89.2 / 82.3 & 90.4 / 86.0 / 62.7 & 89.0 / 87.2 / 85.4 & 82.7 / 86.8 / 91.9 & 92.1 / 80.7 / 76.0 \\
\cline{2-10}
& Average All & 92.7 / 84.6 / 84.4 & 88.8 / 83.9 / 90.6 & 94.4 / 91.3 / 92.6 & 93.9 / 91.4 / 92.1 & 94.9 / 89.8 / 83.9 & 94.6 / 89.6 / 88.7 & 85.5 / 89.3 / 95.0 & 89.6 / 85.4 / 87.6 \\

            \hline
        \end{tabular}
}
\end{table*}
\begin{table*}[htb]
    \centering
    \caption{FUIAD performance~(S-AUROC / I-AUROC / P-AUPRO) comparisons with state-of-the-art anomaly detection methods on multi-view Real-IAD with a noisy ratio of 0.1. 
    }
    \label{tab:mv-fuiad-1}
    \renewcommand{\arraystretch}{1.2}
    \setlength\tabcolsep{6.0pt}
    \resizebox{1.0\linewidth}{!}{
        \begin{tabular}{ll | cc cc | cc | cc}
            \hline
            \multicolumn{2}{c|}{\multirow{2}{*}{Category}} & \multicolumn{4}{c|}{Embedding-based} & \multicolumn{2}{c|}{Data-Aug-based} & \multicolumn{2}{c}{Reconstruction-based} \\ 
            \cline{3-6} \cline{7-8} \cline{9-10}
            & & \textbf{PaDim\cite{padim}} &\textbf{CFlow}\cite{cflow} &\textbf{PatchCore\cite{patchcore}}   &\textbf{SoftPatch\cite{softpatch}} & \textbf{SimpleNet\cite{simplenet}} & \textbf{DeSTSeg\cite{destseg}} & \textbf{RD\cite{rd}} & \textbf{UniAD\cite{uniad}} \\
            \hline
            \multirow{30}{*}{\rotatebox{90}{FUAD-NR-0.1}} & Audiojack & 89.2 / 74.3 / 86.1 & 75.8 / 71.8 / 86.9 & 88.2 / 87.1 / 89.6 & 88.0 / 88.7 / 93.1 & 84.0 / 84.1 / 85.1 & 88.8 / 87.6 / 90.6 & 71.6 / 85.8 / 91.3 & 88.5 / 85.2 / 85.8 \\
& PCB & 84.8 / 76.0 / 71.7 & 79.9 / 74.8 / 86.6 & 93.9 / 93.5 / 94.3 & 91.9 / 93.3 / 95.9 & 84.3 / 83.0 / 87.1 & 87.5 / 85.7 / 87.8 & 90.6 / 91.3 / 94.1 & 85.2 / 83.5 / 83.9 \\
& Phone Battery & 80.3 / 85.4 / 90.0 & 86.7 / 86.1 / 94.6 & 90.1 / 92.5 / 95.8 & 93.4 / 93.0 / 89.0 & 80.9 / 81.8 / 74.7 & 76.5 / 85.6 / 88.2 & 74.4 / 89.2 / 98.0 & 82.7 / 87.6 / 92.8 \\
& SIM Card Set & 88.2 / 92.5 / 93.8 & 88.3 / 90.2 / 93.9 & 94.9 / 97.1 / 97.1 & 96.0 / 96.4 / 91.9 & 89.6 / 85.4 / 62.9 & 81.9 / 82.2 / 87.2 & 67.7 / 91.0 / 96.2 & 90.2 / 91.8 / 87.5 \\
& Switch & 77.5 / 78.3 / 89.7 & 40.4 / 68.3 / 86.3 & 94.2 / 89.5 / 93.4 & 98.3 / 95.6 / 96.7 & 95.5 / 87.9 / 88.4 & 94.4 / 90.1 / 92.4 & 64.2 / 83.4 / 93.4 & 94.8 / 86.8 / 92.2 \\
& Terminal Block & 94.2 / 91.9 / 97.3 & 89.5 / 86.8 / 96.3 & 96.8 / 95.5 / 98.2 & 97.5 / 97.0 / 98.4 & 97.4 / 90.2 / 93.5 & 96.8 / 93.2 / 97.2 & 93.4 / 96.4 / 99.0 & 95.4 / 89.2 / 95.0 \\
& Toothbrush & 74.8 / 72.8 / 79.4 & 77.9 / 78.1 / 86.5 & 87.9 / 89.0 / 93.3 & 90.1 / 90.8 / 89.7 & 82.7 / 82.5 / 72.9 & 88.5 / 82.1 / 72.6 & 74.4 / 81.5 / 91.7 & 92.2 / 83.3 / 88.3 \\
& Toy & 76.5 / 71.3 / 81.3 & 66.7 / 58.6 / 76.7 & 93.1 / 87.9 / 92.2 & 93.4 / 89.1 / 92.7 & 86.1 / 74.7 / 72.2 & 86.3 / 82.7 / 77.7 & 77.1 / 84.7 / 93.5 & 76.5 / 68.0 / 77.8 \\
& Transistor1 & 83.0 / 85.6 / 92.7 & 96.4 / 91.1 / 94.7 & 99.5 / 97.1 / 97.9 & 99.3 / 96.9 / 98.0 & 96.8 / 89.7 / 89.9 & 97.3 / 92.6 / 90.1 & 99.6 / 97.3 / 98.3 & 99.4 / 93.9 / 93.8 \\
& USB & 67.8 / 68.6 / 82.3 & 76.3 / 76.1 / 86.5 & 93.3 / 92.6 / 96.5 & 93.8 / 94.0 / 98.0 & 89.6 / 85.7 / 90.7 & 92.8 / 89.8 / 93.4 & 93.5 / 94.5 / 97.3 & 80.5 / 80.7 / 88.1 \\
& USB Adaptor & 88.6 / 83.9 / 92.6 & 79.4 / 71.1 / 89.8 & 87.8 / 79.4 / 92.8 & 92.6 / 88.1 / 93.2 & 87.8 / 81.1 / 83.7 & 87.7 / 80.5 / 88.1 & 81.2 / 77.5 / 91.8 & 85.2 / 79.0 / 83.7 \\
& Zipper & 97.2 / 89.9 / 81.6 & 89.9 / 88.2 / 94.7 & 99.8 / 97.6 / 91.3 & 99.4 / 97.2 / 91.3 & 96.9 / 91.2 / 83.2 & 90.4 / 88.7 / 85.0 & 99.1 / 98.6 / 98.2 & 97.8 / 97.6 / 96.3 \\
& Bottle Cap & 97.3 / 94.0 / 97.2 & 98.8 / 92.9 / 98.2 & 99.4 / 96.7 / 97.8 & 99.4 / 97.6 / 98.6 & 93.5 / 88.3 / 89.4 & 96.0 / 92.8 / 95.9 & 98.7 / 95.6 / 98.2 & 99.0 / 93.9 / 96.7 \\
& End Cap & 73.3 / 71.9 / 85.4 & 71.0 / 67.1 / 83.1 & 92.7 / 83.5 / 91.8 & 95.0 / 87.6 / 93.6 & 88.3 / 74.7 / 69.1 & 81.5 / 78.2 / 87.9 & 63.7 / 79.1 / 93.3 & 88.8 / 79.0 / 83.8 \\
& Fire Hood & 84.5 / 78.4 / 78.4 & 83.4 / 80.4 / 94.1 & 86.3 / 84.6 / 90.0 & 86.5 / 86.2 / 91.1 & 81.5 / 78.2 / 69.5 & 85.7 / 82.1 / 83.4 & 72.8 / 83.6 / 93.5 & 80.7 / 79.3 / 85.9 \\
& Mounts & 97.7 / 87.8 / 89.2 & 96.5 / 88.8 / 95.6 & 99.8 / 93.1 / 90.2 & 100.0 / 90.6 / 88.7 & 92.6 / 83.8 / 86.3 & 97.6 / 87.6 / 89.4 & 81.7 / 90.3 / 93.6 & 97.5 / 89.9 / 94.3 \\
& Plastic Nut & 84.1 / 80.7 / 91.1 & 84.3 / 82.0 / 94.1 & 97.9 / 93.5 / 96.1 & 96.2 / 93.7 / 96.9 & 87.7 / 85.2 / 76.8 & 91.1 / 90.3 / 95.2 & 90.8 / 89.9 / 97.2 & 87.5 / 80.7 / 90.0 \\
& Plastic Plug & 81.6 / 85.0 / 89.4 & 89.9 / 87.7 / 93.6 & 92.3 / 93.0 / 96.0 & 90.5 / 92.2 / 95.8 & 89.8 / 84.9 / 78.1 & 92.1 / 86.7 / 86.5 & 92.4 / 93.3 / 96.7 & 79.3 / 81.5 / 88.2 \\
& Regulator & 82.7 / 75.4 / 90.1 & 65.0 / 65.2 / 84.3 & 84.4 / 86.1 / 96.9 & 85.8 / 85.1 / 96.8 & 84.4 / 81.3 / 86.5 & 92.2 / 88.1 / 88.1 & 85.6 / 87.2 / 98.0 & 63.3 / 58.9 / 73.1 \\
& Rolled Strip Base & 97.6 / 97.1 / 98.2 & 96.8 / 91.3 / 98.9 & 99.3 / 98.6 / 98.9 & 99.7 / 99.3 / 99.2 & 97.5 / 95.0 / 93.2 & 94.3 / 94.5 / 97.1 & 98.6 / 97.0 / 98.9 & 98.8 / 98.2 / 98.3 \\
& Toy Brick & 85.0 / 71.7 / 78.8 & 72.7 / 72.6 / 89.0 & 84.2 / 81.6 / 85.2 & 81.6 / 81.0 / 87.3 & 72.6 / 66.7 / 56.7 & 84.7 / 76.5 / 76.5 & 73.0 / 72.7 / 88.3 & 80.0 / 76.8 / 83.1 \\
& U Block & 95.8 / 88.8 / 94.7 & 91.2 / 89.3 / 96.8 & 96.4 / 94.8 / 97.7 & 97.0 / 94.1 / 97.3 & 89.3 / 88.4 / 88.1 & 82.5 / 82.3 / 92.6 & 94.1 / 93.6 / 98.3 & 94.2 / 90.5 / 93.6 \\
& Vcpill & 85.4 / 71.1 / 83.9 & 82.5 / 84.3 / 93.7 & 94.5 / 92.0 / 94.6 & 93.3 / 89.9 / 90.5 & 89.6 / 85.3 / 78.5 & 90.8 / 85.6 / 82.2 & 90.1 / 76.1 / 94.9 & 82.2 / 86.3 / 93.9 \\
& Wooden Beads & 89.4 / 88.6 / 87.5 & 89.1 / 89.9 / 91.1 & 85.0 / 87.9 / 89.5 & 88.6 / 89.2 / 90.1 & 83.6 / 82.4 / 77.9 & 82.5 / 81.4 / 76.3 & 57.4 / 82.8 / 91.8 & 79.2 / 79.6 / 84.9 \\
& Woodstick & 79.1 / 82.1 / 79.8 & 73.2 / 73.6 / 89.4 & 73.9 / 80.8 / 70.8 & 74.3 / 79.9 / 72.4 & 74.6 / 74.4 / 71.3 & 82.4 / 87.5 / 88.4 & 81.6 / 88.7 / 93.8 & 76.9 / 81.9 / 78.3 \\
& Tape & 97.3 / 96.7 / 97.4 & 96.5 / 96.7 / 99.2 & 98.6 / 98.3 / 98.1 & 99.5 / 98.7 / 98.0 & 93.7 / 91.5 / 89.5 & 89.9 / 93.8 / 97.9 & 97.0 / 98.2 / 99.4 & 99.2 / 97.9 / 98.6 \\
& Porcelain Doll & 90.6 / 80.5 / 83.7 & 81.5 / 76.9 / 95.9 & 95.9 / 90.7 / 96.7 & 94.8 / 88.5 / 96.1 & 94.0 / 87.1 / 89.1 & 89.2 / 83.2 / 86.0 & 94.0 / 90.4 / 97.7 & 92.5 / 86.3 / 93.5 \\
& Mint & 72.7 / 69.9 / 64.8 & 80.2 / 71.0 / 73.9 & 88.9 / 79.4 / 83.5 & 87.6 / 75.9 / 84.0 & 86.9 / 75.8 / 63.6 & 81.1 / 73.3 / 61.7 & 87.1 / 76.9 / 87.6 & 79.2 / 70.3 / 58.6 \\
& Eraser & 93.6 / 90.3 / 96.0 & 87.0 / 91.1 / 98.0 & 93.9 / 94.5 / 98.4 & 94.8 / 94.8 / 97.8 & 84.3 / 86.7 / 83.8 & 87.3 / 89.0 / 92.9 & 75.5 / 90.1 / 97.7 & 89.6 / 92.0 / 94.7 \\
& Button Battery & 75.8 / 75.7 / 67.4 & 66.7 / 68.4 / 78.4 & 83.1 / 83.8 / 90.1 & 85.1 / 84.2 / 86.4 & 78.5 / 72.9 / 63.8 & 73.8 / 74.6 / 79.8 & 84.0 / 84.9 / 91.7 & 80.3 / 75.2 / 75.5 \\
\cline{2-10}
& Average All & 85.5 / 81.9 / 86.4 & 81.8 / 80.3 / 90.7 & 92.2 / 90.4 / 93.2 & 92.8 / 90.9 / 92.9 & 87.8 / 83.3 / 79.9 & 88.1 / 85.6 / 86.9 & 83.5 / 88.1 / 95.1 & 87.2 / 84.2 / 87.7 \\
            \hline
        \end{tabular}
}
\end{table*}
\begin{table*}[htb]
    \centering
    \caption{FUIAD performance~(S-AUROC / I-AUROC / P-AUPRO) comparisons with state-of-the-art anomaly detection methods on multi-view Real-IAD with a noisy ratio of 0.2. 
    }
    \label{tab:mv-fuiad-2}
    \renewcommand{\arraystretch}{1.2}
    \setlength\tabcolsep{6.0pt}
    \resizebox{1.0\linewidth}{!}{
        \begin{tabular}{ll | cc cc | cc | cc}
            \hline
            \multicolumn{2}{c|}{\multirow{2}{*}{Category}} & \multicolumn{4}{c|}{Embedding-based} & \multicolumn{2}{c|}{Data-Aug-based} & \multicolumn{2}{c}{Reconstruction-based} \\ 
            \cline{3-6} \cline{7-8} \cline{9-10}
            & & \textbf{PaDim\cite{padim}} &\textbf{CFlow}\cite{cflow} &\textbf{PatchCore\cite{patchcore}}   &\textbf{SoftPatch\cite{softpatch}} & \textbf{SimpleNet\cite{simplenet}} & \textbf{DeSTSeg\cite{destseg}} & \textbf{RD\cite{rd}} & \textbf{UniAD\cite{uniad}} \\
            \hline
            \multirow{30}{*}{\rotatebox{90}{FUAD-NR-0.2}} & Audiojack & 86.6 / 73.8 / 86.8 & 66.6 / 67.5 / 87.2 & 84.8 / 85.9 / 88.8 & 86.8 / 87.6 / 92.8 & 76.7 / 80.8 / 81.4 & 80.1 / 84.1 / 87.6 & 78.3 / 85.3 / 90.9 & 81.4 / 83.6 / 85.1 \\
& PCB & 81.7 / 72.4 / 72.0 & 77.6 / 72.5 / 84.0 & 92.3 / 92.5 / 94.3 & 91.6 / 92.9 / 95.7 & 82.0 / 80.4 / 84.8 & 76.9 / 80.1 / 94.5 & 83.6 / 89.0 / 94.4 & 83.6 / 81.9 / 83.8 \\
& Phone Battery & 75.9 / 83.5 / 89.7 & 82.6 / 81.3 / 93.8 & 84.7 / 91.0 / 95.9 & 92.6 / 92.1 / 88.8 & 77.9 / 79.4 / 69.2 & 80.1 / 86.7 / 90.5 & 78.1 / 88.9 / 98.0 & 78.6 / 85.8 / 92.6 \\
& SIM Card Set & 85.8 / 91.0 / 94.2 & 88.9 / 91.3 / 94.2 & 94.5 / 96.8 / 97.2 & 95.6 / 95.7 / 91.4 & 88.1 / 84.5 / 57.9 & 70.9 / 71.6 / 78.2 & 78.9 / 93.6 / 96.6 & 87.0 / 90.0 / 87.3 \\
& Switch & 75.0 / 75.5 / 89.2 & 77.2 / 74.4 / 87.0 & 93.3 / 89.0 / 93.3 & 98.2 / 95.4 / 96.6 & 92.2 / 85.1 / 86.7 & 92.7 / 84.2 / 79.5 & 67.0 / 81.8 / 93.2 & 94.0 / 85.5 / 91.8 \\
& Terminal Block & 92.9 / 90.7 / 97.5 & 86.5 / 83.8 / 96.0 & 96.8 / 95.2 / 98.3 & 98.0 / 97.0 / 98.5 & 90.7 / 85.1 / 88.8 & 89.1 / 86.6 / 97.4 & 94.7 / 96.4 / 99.1 & 95.4 / 89.1 / 95.4 \\
& Toothbrush & 69.3 / 69.8 / 79.0 & 83.4 / 77.9 / 87.1 & 86.6 / 87.5 / 93.2 & 88.9 / 89.5 / 89.5 & 75.2 / 71.6 / 63.6 & 84.1 / 72.8 / 65.0 & 78.9 / 82.3 / 91.1 & 89.3 / 81.7 / 88.2 \\
& Toy & 70.6 / 65.9 / 81.6 & 64.3 / 58.9 / 78.5 & 91.5 / 86.0 / 91.5 & 93.0 / 88.0 / 92.5 & 78.8 / 69.7 / 69.2 & 85.3 / 77.8 / 76.1 & 76.6 / 84.0 / 93.1 & 74.1 / 66.9 / 78.0 \\
& Transistor1 & 76.7 / 82.1 / 92.6 & 96.6 / 91.4 / 95.2 & 99.1 / 96.5 / 97.8 & 97.7 / 95.6 / 97.6 & 90.8 / 82.3 / 81.9 & 93.1 / 83.6 / 74.4 & 99.2 / 95.9 / 98.0 & 99.1 / 92.7 / 93.7 \\
& USB & 62.4 / 66.2 / 82.0 & 77.2 / 77.6 / 87.5 & 92.2 / 92.1 / 96.5 & 93.8 / 93.6 / 98.0 & 81.7 / 82.3 / 86.9 & 82.7 / 84.8 / 89.6 & 93.3 / 94.3 / 97.0 & 80.4 / 80.9 / 87.7 \\
& USB Adaptor & 85.8 / 81.9 / 92.5 & 76.0 / 73.2 / 88.8 & 85.7 / 76.8 / 92.7 & 91.5 / 88.0 / 93.1 & 70.8 / 70.5 / 73.4 & 85.9 / 80.4 / 89.8 & 55.3 / 70.7 / 90.3 & 83.0 / 76.7 / 81.5 \\
& Zipper & 95.2 / 88.1 / 80.9 & 88.4 / 85.3 / 95.3 & 99.6 / 97.3 / 91.4 & 99.0 / 96.9 / 91.4 & 91.0 / 86.0 / 74.6 & 88.1 / 84.6 / 76.8 & 76.5 / 92.3 / 98.1 & 97.0 / 96.9 / 96.2 \\
& Bottle Cap & 96.7 / 93.3 / 97.3 & 96.3 / 93.0 / 98.0 & 98.9 / 96.6 / 98.0 & 99.3 / 97.7 / 98.7 & 87.4 / 84.5 / 86.2 & 91.2 / 91.9 / 97.0 & 92.1 / 93.5 / 98.4 & 98.0 / 93.6 / 96.7 \\
& End Cap & 69.4 / 69.9 / 84.7 & 70.4 / 68.2 / 82.4 & 90.3 / 82.0 / 91.5 & 93.7 / 86.3 / 93.4 & 80.0 / 72.4 / 62.5 & 80.7 / 79.0 / 84.3 & 68.2 / 76.2 / 93.6 & 86.9 / 77.8 / 83.1 \\
& Fire Hood & 80.3 / 76.6 / 78.7 & 85.0 / 84.2 / 95.0 & 83.8 / 83.8 / 89.8 & 85.2 / 85.6 / 91.0 & 75.0 / 76.9 / 76.9 & 86.8 / 81.6 / 85.3 & 73.0 / 82.6 / 93.1 & 79.9 / 78.9 / 85.9 \\
& Mounts & 96.0 / 88.1 / 89.6 & 97.8 / 89.6 / 95.6 & 99.0 / 93.0 / 88.2 & 100.0 / 92.2 / 88.7 & 92.1 / 83.1 / 82.3 & 85.2 / 78.9 / 81.6 & 89.2 / 90.2 / 93.6 & 96.5 / 90.4 / 94.2 \\
& Plastic Nut & 80.0 / 79.5 / 91.5 & 85.4 / 83.3 / 93.5 & 97.4 / 93.3 / 96.2 & 95.5 / 93.1 / 97.0 & 87.4 / 81.1 / 79.0 & 81.3 / 83.5 / 89.3 & 74.7 / 87.0 / 96.8 & 82.8 / 77.2 / 89.0 \\
& Plastic Plug & 79.5 / 83.7 / 89.8 & 69.4 / 75.3 / 94.4 & 91.8 / 92.6 / 96.5 & 89.9 / 92.1 / 96.0 & 85.7 / 83.0 / 74.2 & 84.8 / 81.4 / 70.4 & 74.1 / 89.7 / 96.8 & 78.1 / 78.9 / 87.4 \\
& Regulator & 78.5 / 73.5 / 90.5 & 70.9 / 67.6 / 85.9 & 85.0 / 85.1 / 97.2 & 85.1 / 85.4 / 97.1 & 80.3 / 74.3 / 81.0 & 87.9 / 84.0 / 84.8 & 82.9 / 86.8 / 98.1 & 58.9 / 58.3 / 73.5 \\
& Rolled Strip Base & 96.4 / 96.2 / 98.1 & 94.2 / 90.1 / 98.9 & 98.6 / 97.6 / 98.8 & 99.5 / 98.8 / 99.1 & 96.2 / 91.5 / 86.7 & 87.4 / 89.4 / 94.3 & 93.7 / 93.7 / 99.0 & 98.7 / 97.9 / 98.3 \\
& Toy Brick & 82.9 / 70.4 / 78.7 & 76.6 / 76.1 / 88.6 & 83.4 / 80.1 / 85.0 & 80.8 / 80.5 / 87.1 & 67.6 / 65.7 / 58.7 & 64.1 / 64.9 / 70.8 & 84.1 / 78.5 / 88.5 & 76.7 / 75.4 / 81.8 \\
& U Block & 94.5 / 87.9 / 95.0 & 87.5 / 88.3 / 96.0 & 95.8 / 94.7 / 97.8 & 96.5 / 94.0 / 97.6 & 84.0 / 83.2 / 78.6 & 82.9 / 80.5 / 89.6 & 94.6 / 93.5 / 98.3 & 93.7 / 89.9 / 93.4 \\
& Vcpill & 77.5 / 68.3 / 83.9 & 83.1 / 84.9 / 94.7 & 93.3 / 90.9 / 94.4 & 92.8 / 89.4 / 90.5 & 83.8 / 82.7 / 78.1 & 86.6 / 81.4 / 75.6 & 81.8 / 91.3 / 96.0 & 81.4 / 86.7 / 94.0 \\
& Wooden Beads & 85.7 / 87.2 / 88.4 & 85.7 / 84.3 / 92.1 & 84.0 / 87.4 / 89.3 & 85.0 / 88.5 / 90.0 & 81.5 / 80.7 / 76.8 & 77.1 / 78.0 / 79.8 & 84.5 / 87.4 / 91.6 & 79.1 / 79.0 / 84.1 \\
& Woodstick & 77.4 / 80.8 / 79.3 & 67.7 / 74.9 / 87.0 & 73.2 / 79.2 / 69.7 & 75.4 / 80.1 / 72.7 & 72.8 / 75.2 / 69.6 & 79.7 / 85.7 / 92.7 & 63.1 / 83.8 / 92.5 & 77.6 / 80.7 / 77.3 \\
& Tape & 96.7 / 96.2 / 97.3 & 98.0 / 95.6 / 99.3 & 98.4 / 97.9 / 98.0 & 99.7 / 98.6 / 98.0 & 86.4 / 89.3 / 86.9 & 83.0 / 89.8 / 94.8 & 94.0 / 97.4 / 99.3 & 98.9 / 97.8 / 98.6 \\
& Porcelain Doll & 85.6 / 79.7 / 84.1 & 75.5 / 76.8 / 96.1 & 95.6 / 90.3 / 96.6 & 95.0 / 87.9 / 96.2 & 83.3 / 81.2 / 78.1 & 85.6 / 83.2 / 92.8 & 91.2 / 89.6 / 97.7 & 91.5 / 86.2 / 93.4 \\
& Mint & 69.8 / 68.9 / 65.0 & 79.4 / 68.1 / 72.8 & 88.4 / 78.3 / 82.9 & 87.2 / 75.5 / 84.2 & 81.0 / 71.7 / 52.9 & 52.0 / 54.7 / 43.7 & 68.0 / 72.7 / 86.0 & 72.0 / 64.7 / 57.2 \\
& Eraser & 93.0 / 90.1 / 96.6 & 78.4 / 83.9 / 98.0 & 92.5 / 94.4 / 98.4 & 94.5 / 94.1 / 97.8 & 84.2 / 84.6 / 87.9 & 82.5 / 85.3 / 85.0 & 76.3 / 90.0 / 97.7 & 86.9 / 91.0 / 94.7 \\
& Button Battery & 68.7 / 71.5 / 68.7 & 73.2 / 67.1 / 77.9 & 81.8 / 82.6 / 89.5 & 84.1 / 82.9 / 86.0 & 77.8 / 69.7 / 62.7 & 54.2 / 58.4 / 86.1 & 74.6 / 80.5 / 90.8 & 75.4 / 67.2 / 74.8 \\
\cline{2-10}
& Average All & 82.2 / 80.1 / 86.5 & 81.3 / 79.6 / 90.7 & 91.1 / 89.5 / 93.0 & 92.2 / 90.5 / 92.9 & 82.7 / 79.6 / 75.9 & 81.4 / 80.3 / 83.2 & 80.7 / 87.3 / 94.9 & 85.2 / 82.8 / 87.3 \\
            \hline
        \end{tabular}
}
\end{table*}
\begin{table*}[htb]
    \centering
    \caption{FUIAD performance~(S-AUROC / I-AUROC / P-AUPRO) comparisons with state-of-the-art anomaly detection methods on multi-view Real-IAD with a noisy ratio of 0.4. 
    }
    \label{tab:mv-fuiad-4}
    \renewcommand{\arraystretch}{1.2}
    \setlength\tabcolsep{6.0pt}
    \resizebox{1.0\linewidth}{!}{
        \begin{tabular}{ll | cc cc | cc | cc}
            \hline
            \multicolumn{2}{c|}{\multirow{2}{*}{Category}} & \multicolumn{4}{c|}{Embedding-based} & \multicolumn{2}{c|}{Data-Aug-based} & \multicolumn{2}{c}{Reconstruction-based} \\ 
            \cline{3-6} \cline{7-8} \cline{9-10}
            & & \textbf{PaDim\cite{padim}} &\textbf{CFlow}\cite{cflow} &\textbf{PatchCore\cite{patchcore}}   &\textbf{SoftPatch\cite{softpatch}} & \textbf{SimpleNet\cite{simplenet}} & \textbf{DeSTSeg\cite{destseg}} & \textbf{RD\cite{rd}} & \textbf{UniAD\cite{uniad}} \\
            \hline
            \multirow{30}{*}{\rotatebox{90}{FUAD-NR-0.4}} & Audiojack & 80.5 / 70.7 / 86.8 & 66.2 / 73.3 / 85.5 & 79.7 / 82.8 / 88.3 & 80.4 / 85.2 / 92.7 & 73.1 / 73.3 / 82.9 & 76.9 / 83.4 / 74.8 & 56.4 / 77.3 / 90.4 & 78.1 / 80.4 / 84.5 \\
& PCB & 74.0 / 66.7 / 71.4 & 70.1 / 70.6 / 86.3 & 90.7 / 91.2 / 93.9 & 91.1 / 91.4 / 95.6 & 80.8 / 76.0 / 84.3 & 73.2 / 72.5 / 82.1 & 87.0 / 89.1 / 94.1 & 77.3 / 75.7 / 82.3 \\
& Phone Battery & 69.4 / 80.5 / 89.1 & 73.5 / 78.9 / 94.0 & 80.2 / 88.4 / 95.1 & 87.9 / 91.0 / 88.7 & 67.2 / 73.5 / 55.3 & 62.0 / 78.3 / 83.0 & 70.7 / 86.8 / 97.7 & 72.0 / 82.0 / 91.6 \\
& SIM Card Set & 78.7 / 87.2 / 93.9 & 86.5 / 88.0 / 94.1 & 92.8 / 96.1 / 97.4 & 94.3 / 95.0 / 90.9 & 82.2 / 74.9 / 52.4 & 78.3 / 80.2 / 86.6 & 42.2 / 83.9 / 96.7 & 80.7 / 84.7 / 86.4 \\
& Switch & 68.0 / 70.2 / 87.9 & 62.4 / 68.6 / 86.3 & 89.7 / 87.1 / 91.3 & 97.8 / 94.8 / 95.1 & 80.6 / 76.3 / 75.5 & 89.6 / 85.0 / 84.8 & 73.9 / 71.8 / 90.6 & 88.8 / 82.6 / 90.8 \\
& Terminal Block & 88.0 / 88.2 / 97.7 & 79.7 / 77.5 / 95.3 & 95.3 / 94.1 / 98.4 & 97.2 / 96.8 / 98.6 & 83.8 / 79.2 / 84.1 & 44.7 / 71.9 / 92.9 & 84.8 / 92.6 / 99.1 & 91.5 / 87.2 / 95.2 \\
& Toothbrush & 63.2 / 66.3 / 78.2 & 67.3 / 66.9 / 86.2 & 84.9 / 85.6 / 92.8 & 87.3 / 87.6 / 89.2 & 67.7 / 66.1 / 56.4 & 43.8 / 59.5 / 69.6 & 81.3 / 81.3 / 91.0 & 82.3 / 76.9 / 87.0 \\
& Toy & 63.1 / 58.7 / 80.6 & 55.3 / 52.9 / 75.3 & 90.2 / 83.5 / 91.2 & 91.1 / 86.5 / 92.1 & 75.8 / 65.1 / 58.6 & 77.3 / 68.3 / 75.5 & 64.8 / 76.2 / 92.3 & 60.9 / 58.8 / 74.8 \\
& Transistor1 & 66.2 / 76.0 / 91.6 & 91.5 / 87.0 / 94.7 & 98.5 / 94.5 / 97.4 & 95.8 / 92.8 / 96.5 & 83.2 / 76.1 / 79.4 & 70.4 / 68.9 / 47.2 & 95.4 / 92.9 / 97.8 & 96.0 / 87.9 / 92.7 \\
& USB & 50.4 / 59.9 / 80.1 & 75.1 / 73.7 / 83.6 & 88.6 / 89.9 / 95.3 & 93.2 / 93.3 / 97.8 & 73.9 / 78.2 / 81.7 & 84.3 / 83.1 / 84.5 & 77.0 / 89.4 / 96.4 & 77.9 / 76.9 / 87.2 \\
& USB Adaptor & 82.9 / 79.7 / 92.1 & 72.7 / 66.9 / 88.7 & 82.9 / 74.4 / 92.0 & 90.3 / 85.1 / 92.9 & 69.3 / 67.7 / 61.7 & 81.1 / 71.5 / 82.3 & 79.8 / 73.1 / 91.1 & 75.4 / 71.0 / 79.6 \\
& Zipper & 91.9 / 85.6 / 80.0 & 85.5 / 81.7 / 94.7 & 98.9 / 96.6 / 91.2 & 98.1 / 96.3 / 91.1 & 90.2 / 84.6 / 77.6 & 81.5 / 78.4 / 68.8 & 97.0 / 96.7 / 98.0 & 96.0 / 94.9 / 95.5 \\
& Bottle Cap & 94.9 / 90.9 / 97.1 & 89.8 / 85.5 / 97.9 & 97.8 / 95.5 / 98.0 & 98.7 / 97.4 / 98.7 & 87.8 / 79.0 / 81.6 & 83.2 / 79.3 / 75.5 & 84.4 / 92.2 / 98.3 & 97.5 / 92.6 / 96.8 \\
& End Cap & 63.9 / 67.3 / 84.0 & 78.8 / 71.5 / 83.5 & 87.6 / 79.8 / 91.0 & 90.0 / 84.5 / 92.2 & 72.2 / 68.3 / 58.0 & 78.2 / 76.8 / 70.2 & 81.1 / 77.0 / 92.8 & 84.4 / 76.2 / 83.3 \\
& Fire Hood & 74.7 / 74.9 / 78.4 & 84.9 / 82.7 / 94.4 & 81.0 / 83.0 / 89.6 & 83.7 / 84.9 / 91.0 & 77.1 / 73.5 / 70.6 & 82.8 / 79.0 / 80.7 & 48.4 / 78.0 / 93.1 & 78.5 / 77.9 / 85.8 \\
& Mounts & 93.8 / 86.9 / 90.0 & 96.2 / 88.7 / 96.3 & 98.2 / 92.6 / 85.8 & 98.3 / 91.9 / 89.4 & 82.1 / 79.1 / 69.7 & 83.2 / 76.3 / 72.3 & 62.0 / 75.8 / 94.3 & 94.0 / 90.5 / 94.2 \\
& Plastic Nut & 71.7 / 77.1 / 91.5 & 78.2 / 82.2 / 94.2 & 95.8 / 92.6 / 96.0 & 94.6 / 91.6 / 96.9 & 74.9 / 77.8 / 69.9 & 76.7 / 75.2 / 61.4 & 86.3 / 88.6 / 96.6 & 79.5 / 74.5 / 88.2 \\
& Plastic Plug & 77.5 / 82.0 / 90.1 & 87.1 / 86.3 / 94.2 & 89.6 / 92.2 / 96.6 & 89.2 / 91.7 / 95.8 & 85.3 / 79.9 / 70.0 & 85.2 / 82.3 / 83.3 & 83.1 / 89.7 / 96.7 & 72.7 / 75.2 / 87.3 \\
& Regulator & 72.6 / 70.6 / 90.4 & 67.9 / 63.1 / 82.1 & 82.6 / 84.3 / 96.7 & 83.7 / 84.9 / 96.8 & 73.2 / 71.5 / 78.3 & 82.6 / 80.7 / 86.5 & 74.8 / 83.8 / 97.8 & 59.5 / 57.1 / 72.7 \\
& Rolled Strip Base & 94.4 / 94.4 / 98.0 & 91.6 / 91.1 / 98.8 & 97.3 / 95.8 / 98.7 & 98.3 / 96.6 / 98.8 & 91.2 / 83.7 / 66.4 & 81.7 / 80.2 / 86.3 & 88.7 / 91.0 / 98.6 & 98.0 / 96.5 / 98.2 \\
& Toy Brick & 78.4 / 68.3 / 78.2 & 76.0 / 75.9 / 87.1 & 80.2 / 78.1 / 84.3 & 80.4 / 78.0 / 86.4 & 66.6 / 63.8 / 65.7 & 63.1 / 60.8 / 62.3 & 81.7 / 78.9 / 88.5 & 73.9 / 73.1 / 80.9 \\
& U Block & 91.7 / 87.2 / 95.2 & 90.0 / 89.3 / 96.5 & 94.8 / 94.5 / 97.8 & 95.8 / 93.4 / 97.5 & 77.0 / 78.5 / 73.9 & 80.3 / 81.5 / 87.0 & 91.9 / 93.0 / 98.2 & 90.5 / 88.1 / 93.0 \\
& Vcpill & 63.7 / 64.8 / 83.3 & 79.9 / 84.1 / 93.8 & 90.3 / 89.4 / 93.9 & 90.2 / 88.6 / 90.0 & 82.6 / 77.8 / 71.2 & 66.8 / 68.6 / 61.2 & 69.1 / 84.5 / 94.2 & 73.4 / 83.5 / 93.2 \\
& Wooden Beads & 81.6 / 84.7 / 88.5 & 84.6 / 86.2 / 92.7 & 81.5 / 85.9 / 88.8 & 82.7 / 87.6 / 89.2 & 75.9 / 77.3 / 72.3 & 57.1 / 51.2 / 66.1 & 73.8 / 82.2 / 91.2 & 73.9 / 75.4 / 84.0 \\
& Woodstick & 73.6 / 79.0 / 78.7 & 67.9 / 73.6 / 87.1 & 70.8 / 78.1 / 68.3 & 74.7 / 79.7 / 71.8 & 69.8 / 71.0 / 67.2 & 76.8 / 79.0 / 73.3 & 46.1 / 79.9 / 92.6 & 76.1 / 79.4 / 77.6 \\
& Tape & 95.5 / 95.4 / 97.3 & 96.3 / 95.9 / 99.3 & 98.0 / 97.7 / 98.0 & 99.2 / 97.9 / 97.9 & 82.6 / 83.4 / 80.6 & 85.8 / 74.8 / 88.9 & 98.3 / 97.8 / 99.3 & 97.0 / 97.0 / 98.4 \\
& Porcelain Doll & 76.5 / 76.4 / 83.7 & 74.7 / 75.6 / 95.3 & 94.3 / 89.0 / 96.4 & 93.5 / 86.7 / 95.9 & 86.7 / 80.5 / 81.4 & 78.8 / 76.9 / 77.6 & 81.6 / 87.8 / 97.8 & 87.1 / 83.0 / 93.1 \\
& Mint & 63.5 / 66.2 / 64.6 & 73.2 / 69.6 / 71.7 & 86.3 / 77.5 / 82.3 & 84.9 / 74.1 / 83.2 & 77.4 / 67.0 / 56.8 & 72.3 / 67.6 / 64.4 & 78.2 / 75.2 / 86.9 & 74.1 / 66.8 / 56.3 \\
& Eraser & 89.3 / 88.8 / 96.9 & 87.4 / 89.4 / 97.7 & 89.9 / 92.8 / 98.3 & 93.6 / 94.0 / 97.7 & 72.7 / 77.2 / 75.9 & 73.9 / 77.7 / 80.3 & 83.4 / 89.7 / 97.8 & 85.4 / 89.6 / 94.3 \\
& Button Battery & 59.9 / 64.3 / 68.9 & 67.5 / 64.0 / 79.8 & 79.7 / 79.9 / 88.3 & 82.7 / 80.7 / 85.0 & 68.3 / 62.3 / 56.2 & 64.4 / 62.2 / 56.2 & 79.7 / 80.0 / 90.4 & 72.8 / 66.5 / 73.9 \\
\cline{2-10}
& Average All & 76.4 / 77.0 / 86.1 & 78.6 / 78.0 / 90.2 & 88.9 / 88.1 / 92.4 & 90.6 / 89.3 / 92.5 & 77.7 / 74.7 / 70.5 & 74.5 / 74.4 / 75.5 & 76.8 / 84.5 / 94.7 & 81.5 / 80.1 / 86.6 \\
            \hline
        \end{tabular}
}
\end{table*}

\end{document}